\documentclass[10pt,twocolumn,letterpaper]{article}
\usepackage{cvpr}
\usepackage{times}
\usepackage{epsfig}
\usepackage{graphicx}
\usepackage{amsmath}
\usepackage{amssymb}
\newtheorem{theorem}{Theorem}
\usepackage{rotating}
\usepackage{multirow}
\usepackage{pifont}

\usepackage{soul}

 \newcommand{\cmark}{\ding{51}}%
 \newcommand{\xmark}{\ding{55}}%

\usepackage[pagebackref=true,breaklinks=true,letterpaper=true,colorlinks,bookmarks=false]{hyperref}

\cvprfinalcopy 

\def\cvprPaperID{2437} 

\ifcvprfinal\fi
\begin{document}
	
\title{Geometry-aware Deep Network for Single-Image Novel View Synthesis\vspace{-0.2cm}} 
\author{Miaomiao Liu\\
CECS, ANU\\
Canberra, Australia\\
{\tt\small miaomiao.liu@anu.edu.au}
\and
Xuming He\\
ShanghaiTech University\\
Shanghai, China\\
{\tt\small hexm@shanghaitech.edu.cn}
\and
Mathieu Salzmann\\
CVLAB, EPFL\\
Lausanne, Switzerland\\
{\tt\small mathieu.salzmann@epfl.ch}
}
\maketitle


\begin{abstract}
\vspace{-0.3cm}
This paper tackles the problem of novel view synthesis from a single image.~In particular, we target real-world scenes with rich geometric structure, a challenging task due to the large appearance variations of such scenes and the lack of simple 3D models to represent them. Modern, learning-based approaches mostly focus on appearance to synthesize novel views and thus tend to generate predictions that are inconsistent with the underlying scene structure.

By contrast, in this paper, we propose to exploit the 3D geometry of the scene to synthesize a novel view. Specifically, we approximate a real-world scene by a fixed number of planes, and learn to predict a set of homographies and their corresponding region masks to transform the input image into a novel view. To this end, we develop a new region-aware geometric transform network that performs these multiple tasks in a common framework.
Our results on the outdoor KITTI and the indoor ScanNet datasets demonstrate the effectiveness of our network in generating high-quality synthetic views that respect the scene geometry, thus outperforming the state-of-the-art methods. 

\end{abstract}
\vspace{-0.3cm}
\section{Introduction}
\vspace{-0.1cm}

Human beings can easily hallucinate what a scene would look like from a different viewpoint, or, for a dynamic scene, in the near future. Automatically performing such a~\emph{novel view synthesis}, however, remains a challenging task for computer vision systems.

Over the past two decades, the most popular approach to synthesizing new views has been to reconstruct an exact or approximate 3D scene model from multiple views~\cite{zhou2013plane,liu2009content,mcmillan1995plenoptic, woodford2007new,chaurasia2013depth}. By contrast, view synthesis from a single image, which can be applied to a broader range of problems, has received much less attention. To overcome the lack of depth information, early methods have proposed to leverage semantic-based priors~\cite{hoiem2005automatic} and geometric cues, such as vanishing points~\cite{horry1997tour}, which, while effective, tend to be less robust than their multi-view counterparts.

\begin{figure}[t!]
	\vspace{-0.3cm}
	\begin{tabular}{cc}
		Input view & \hspace{-0.2cm}Ground-truth novel view\\
		\includegraphics[width=0.45\linewidth,height=0.3\linewidth]{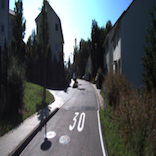}& \hspace{-0.2cm}\includegraphics[width=0.45\linewidth,height=0.3\linewidth]{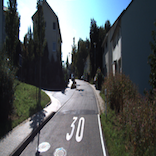} \vspace{0.1cm}\\
		\includegraphics[width=0.45\linewidth,height=0.3\linewidth]{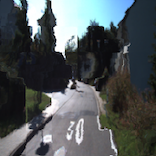}&
		\hspace{-0.2cm}\includegraphics[width=0.45\linewidth,height=0.3\linewidth]{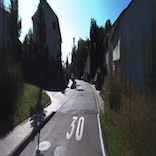}\\
		~\cite{zhou2016view} & Ours\\
	\end{tabular}
	\caption{{\bf Novel view synthesis from a single image.} Given an input image of the scene and a relative pose, we seek to predict a new image of the scene observed from this new viewpoint. To this end, and in contrast with state-of-the-art methods, we propose to explicitly rely on 3D geometry within a deep learning paradigm. As a consequence, and as evidenced by our results, our predictions better respect the scene structure and are thus more realistic.
	}
	\label{fig:intro}
	\vspace{-0.4cm}
\end{figure}

Inspired by the recent deep learning revolution in computer vision, several works have proposed to exploit Deep Convolutional Neural Networks~(CNNs) to tackle the novel view synthesis problem~\cite{flynn2016deepstereo,tatarchenko2016multi,zhou2016view}. Whether predicting image pixels directly~\cite{tatarchenko2016multi}, plane-sweep volumes~\cite{flynn2016deepstereo}, appearance flow~\cite{zhou2016view},  or appearance flow, visibility and the intensity of pixels that were not in the input view~\cite{park2017transformation},
these methods,
in essence, all aim to solely leverage appearance to predict the flow of each pixel from the input view to the novel view without exploiting the flow of the other pixels.
As such, as shown in Fig.~\ref{fig:intro}, they tend to generate artefacts, such as distorted local structures in the synthesized images.

In this paper, we propose to explicitly account for 3D geometry, and thus respect 3D scene structure, in the single-image novel view synthesis process. To this end, we approximate the scene by a fixed number of planes, and learn to predict corresponding homographies that, once applied to the input image, generate a set of candidate images for the novel view. We then learn to predict a selection map 
corresponding to each homography, which, after warping, is used to combine the candidate images to generate the novel view. In essence, our homography-based approach enforces geometric constraints on the flow field, thus modeling scene structure.
Our approach can be thought of as a divide-and-conquer strategy that allows us to encode a 3D geometric prior while learning the image transformation.



To achieve this, we develop a novel deep architecture consisting of two subnetworks.  The first one estimates pixel-wise depth and normals in the input image, which, in conjunction with the relative pose between the input and novel views, are then used to estimate one homography for each planar region in the scene. These homographies then let us produce a set of warped input images. The second subnetwork aims to predict a pixel-wise probability, or selection map encoding to which homography each input pixel should be associated. These maps are then warped with the corresponding predicted homographies, and the novel view is generated by combining the warped input images according to the warped selection maps. To account for pixels not in the input view and potential blur arising from the combination of multiple warped images, inspired by~\cite{park2017transformation}, we further propose to refine the synthesized image with an encoder-decoder network with skip connections. As evidenced by Fig.~\ref{fig:intro}, our complete framework yields realistic-looking novel views.

We demonstrate the effectiveness of our approach on the challenging KITTI odometry dataset~\cite{Geiger2012CVPR} and ScanNet~\cite{dai2017scannet}, depicting complex urban outdoor scenes and indoor scenes, respectively.
Thanks to our geometry-based reasoning, our method not only outperforms the state-of-the-art \emph{appearance~flow} technique of~\cite{zhou2016view} quantitatively, 
but
also yields visually more realistic predictions.

\vspace{-0.15cm}
\section{Related Work}
\label{sec:related}
\vspace{-0.2cm}

Over the years, two main classes of methods have been proposed to address the novel view synthesis problem:  those that rely on geometry, and the more recent ones that exploit deep learning. Below, we review the methods belonging to these two classes.

\vspace{-0.5cm}
\paragraph{Geometry-based view synthesis.} Originally, the most popular approach to view synthesis consisted of explicitly modeling 3D information, via either a detailed 3D model, or an approximate representation of the 3D scene structure. This idea was introduced in~\cite{mcmillan1995plenoptic} more than two decades ago, by relying on multi-view stereo and a warping strategy. With the impressive progress of multi-view 3D reconstruction techniques~\cite{furukawa2010accurate}, highly detailed models can be obtained, and novel views generated by making use of the target pose given as input.  In complex scenes, however, this process remains challenging due to, e.g., occlusions leading to holes in the 3D models. In this context,~\cite{chaurasia2013depth} first reconstructs a partial scene from multiple images, and then synthesizes depth to fill in the missing pixels and correct the unreliable regions. Instead of relying on dense reconstruction,~\cite{zhou2013plane} leverages sparse points obtained from structure-from-motion in conjunction with segmented image regions, each of which is assumed to be planar and associated to a homography to warp the input image. While effective in their context, these methods are inapplicable to the scenario where a single image is available to synthesize a novel view.

Only little work has been done to leverage geometry for single-image novel view synthesis. In particular,~\cite{horry1997tour} models the scene as an axis-aligned box, and requires a user to annotate the box coordinates, vanishing points and foreground to be able to render the model from a different viewpoint. In~\cite{hoiem2005automatic}, the image is labeled into three geometric classes, which defines an approximate scene structure that can be rendered from a new viewpoint. These methods, however, only model a very coarse structure of the scene, and therefore cannot yield realistic novel views.  By contrast, the recent work of~\cite{rematas2016novel} leverages a large collection of 3D models to infer the one closest to an input image. While effective for individual objects, this approach does not translate well to complex, real-world scenes with rich structures and dynamic motion, such as urban ones.

\begin{figure*}[t!]
\vspace{-0.3cm}
	\centering
	\includegraphics[width=0.9\linewidth]{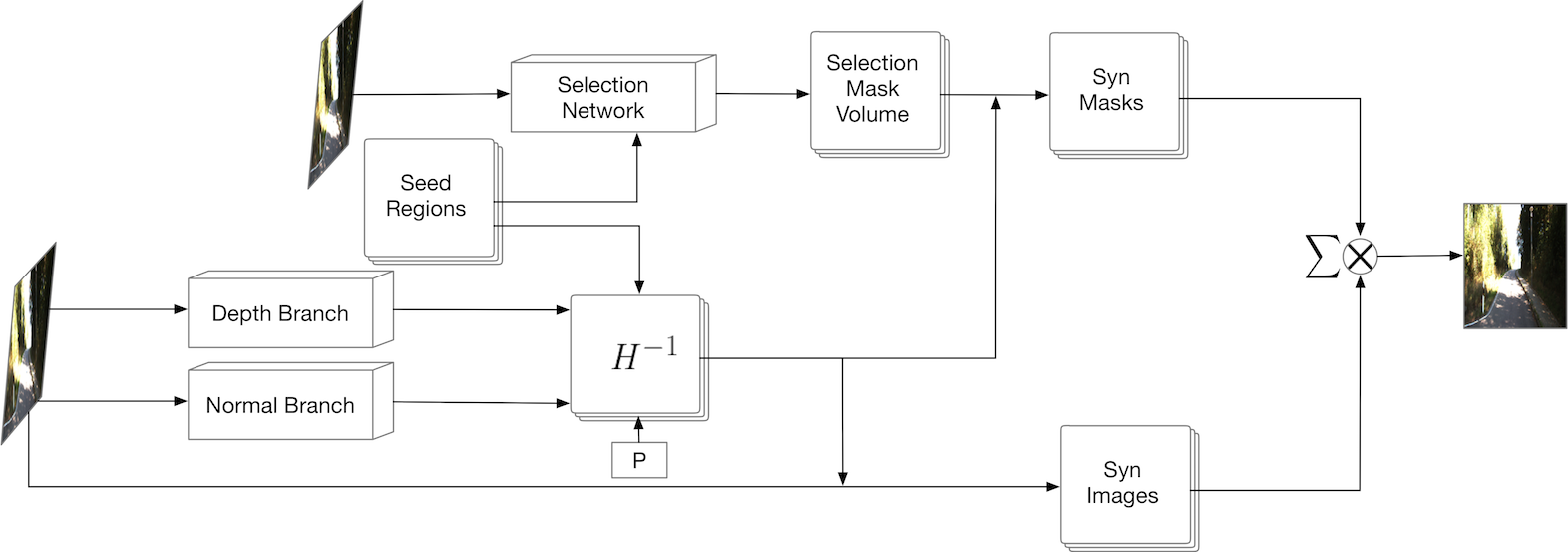} 
	\caption{{\bf Our region-aware geometric-transform network.} To tackle the single-image novel view synthesis problem, we develop a geometry-aware deep architecture consisting of two subnetworks. Given an input image, the first one predicts pixel-wise depth and normal maps. These predictions are then used in conjunction with segmentation masks obtained from the image and the desired relative pose to generate a fixed number of homographies, which are in turn employed to produce warped images. The second subnetwork predicts pixel-wise selection maps that associate each input pixel with one homography. These maps are warped by their respective homographies, and the novel view is obtained by combining the warped images according to the warped selection maps.
	}
	\label{fig:networkStru}
	\vspace{-0.3cm}
\end{figure*}

\vspace{-0.5cm}
\paragraph{View synthesis from CNNs.} With the advent of deep learning in computer vision, CNNs have recently been investigated to generate novel views. In particular,~\cite{flynn2016deepstereo} proposes to synthesize the novel image from neighboring views. To this end, a plane-sweep volume, encoding a set of possible image appearances, was used as input to a network whose goal was to select the correct pixel appearance in the volume. This framework, however, requires a large memory and was only evaluated for view interpolation. Similarly,~\cite{ji2017deep} tackles the view interpolation task from a pair of images, but aims to learn to rectify the two images and predict pixels correspondences. The novel view is generated by fusing the pixels of the image pair using the estimated correspondence. In contrast to these methods, we focus on single-image view synthesis. 

In this context,~\cite{kulkarni2015deep} trains a variational auto-encoder to decouple the image into hidden factors, constrained to correspond to viewpoint and lighting conditions. While this network can generate an image from a new viewpoint by manipulating the hidden factors, it is mostly restricted to small rotations. In~\cite{tatarchenko2016multi}, an encoder-decoder network is trained to directly synthesize the pixels of the new view from the input image and the relative pose. While this network was shown to handle large rotations, the predicted images are typically blurry. Instead of directly synthesizing the image,~\cite{zhou2016view} proposes to predict the displacements of the pixels from the input view to the new one, named the~\emph{appearance flow}. While this method yields sharper results, by predicting the displacements in a pixel-wise manner, it doesn't account for the scene structure, and thus, as illustrated in Fig.~\ref{fig:intro}, introduces unrealistic artefacts. The recent work of~\cite{park2017transformation} builds upon appearance flow by additionally predicting a visibility map, whose goal is to reflect the visibility constraints arising from a 3D object shape. During training, the ground-truth visibility maps are obtained by making use of 3D CAD models of the objects of interest. While this indeed exploits 3D geometry, at test time, the synthesis process neither explicitly encodes notions of geometry nor preserves local geometric structures in the new image. Furthermore, its use of 3D CAD models makes this approach better-suited to single-object view synthesis than to tackling complex real-world scenes.

By contrast, here, we explicitly leverage 3D geometry during the synthesis of the novel view, by developing a deep learning framework that exploits the notion of local homographies. 
As illustrated by Fig.~\ref{fig:intro}, our geometry-aware deep learning strategy yields realistic predictions that better reflect the scene structure.

Note that some work has focused on the specific case of stereo view synthesis, that is, generating an image of one view from that of the other in a stereo setup~\cite{xie2016deep3d}. While effective, this does not generalize to arbitrary novel views, since not all 3D information can be explained by disparity. 
Furthermore, view synthesis has been employed as supervision for depth estimation~\cite{garg2016unsupervised,zhou2017unsupervised}. However, novel views generated from predicted depth maps are typically highly incomplete, and, while suitable for depth estimation, not realistic-looking. Here, we focus on synthesizing realistic novel views with general pose variations.

\vspace{-0.1cm}
\section{Our Approach}
\vspace{-0.1cm}

Our goal is to explicitly leverage information about the 3D scene structure to perform single-image novel view synthesis. To this end, we assume that the scene can be represented with multiple planes and learn to predict their respective homographies, which let us generate a set of candidate images in the new view. We additionally learn to estimate selection maps corresponding to the homographies, which encode to which homography each input pixel should be associated. Warping these maps and using them in conjunction with the candidate new view images lets us synthesize the novel view. We then complete the regions that were unseen in the input view, and thus cannot be synthesized with this strategy, using an encoder-decoder network similar to the generator of~\cite{park2017transformation}. Below, we first introduce our region-aware geometric-transform network, and then discuss this encoder-decoder refinement.

\subsection{Region-aware Geometric-transform Network}
\vspace{-0.1cm}
To learn to predict a novel view from a single image while exploiting the 3D geometry of the scene, we develop the network shown in Fig.~\ref{fig:networkStru}. This architecture consists of two subnetworks. The bottom one first predicts pixel-wise depth and normals from a single image in two independent streams. These predictions are then used, together with region masks extracted from the input image and the relative pose between the input view and the novel one, to compute multiple homographies, which we employ to warp the input image, thus generating candidate synthesized views. The second subnetwork, at the top of Fig.~\ref{fig:networkStru}, predicts selection masks indicating, for each pixel, to which homography it should be associated. We then compute the novel view by assembling the candidate synthesized images according to the warped selection masks. Below, we describe these different stages in more detail.


\vspace{-0.4cm}
\paragraph{Depth and Normal Prediction.}
We use standard fully-convolutional architectures to predict pixel-wise depth and normal maps separately. The details of these architectures are provided in the experiments section.


\vspace{-0.4cm}
\paragraph{Generating Homographies.}
Since we represent the scene as a set of $m$ planar surfaces, a novel view can be obtained by applying one homography to each surface. For one plane, a homography can be computed from its  depth and normal, given the desired relative pose, i.e., 3D rotation and translation, and camera intrinsic parameters. To model $m$ different planes, we make use of a segmentation of the input image into $m$ regions, referred to as \emph{seed regions} and described in Section~\ref{sec:seg}, to pool the above-mentioned pixel-wise depth and normal estimates.

More specifically, let $M$ be an $h\times w \times m$ binary tensor encoding $m$ segmentation masks obtained from the $h \times w$ input image $I^s$. Furthermore, let us denote by $M_j$ the binary mask corresponding to the $j^{th}$ segment. Assuming that each segment is planar, we approximate its normal as 
\begin{equation}
\bar{{\bf n}}_j  = \frac{\sum_{{\bf x} \in \Omega}M_j({\bf x})\cdot {\bf n}({\bf x})}{\sum_{{\bf x} \in \Omega} M_j({\bf x})}\;,
\end{equation}
where $\Omega$ denotes the set of all pixel locations, and ${\bf n}({\bf x})$ corresponds to the normal estimate at location ${\bf x}$. We then normalize $\bar{{\bf n}}_j$ to have unit norm.

A plane with normal $\bar{{\bf n}}_j$ can be defined by a vector $<\bar{n}_j^x,\bar{n}_j^y,\bar{n}_j^z,\bar{n}_j^d>$, such that any 3D point ${\bf Q}$ on the plane satisfies $\bar{{\bf n}}_j^T {\bf Q} + \bar{n}_j^d = 0$. While our average normal estimate provides us with the first 3 parameters, we still need to compute $\bar{n}_j^d$. To this end, let us consider the center of region $j$, with coordinates $(c_j^x, c_j^y)$. We approximate the depth at the center location as
\begin{equation}
\bar{d}_j  = \frac{\sum_{{\bf x} \in \Omega} M_j({\bf x})\cdot d({\bf x})}{\sum_{{\bf x} \in \Omega} M_j({\bf x})}\;,
\end{equation}
where $d({\bf x})$ corresponds to the depth estimate at location ${\bf x}.$ This allows us to increase robustness to noise in the predicted depth map compared to directly using $d(c_j^x,c_j^y)$.
%
Given the matrix of camera intrinsic parameters ${\bf K}$, the corresponding 3D point can be expressed as 
\begin{equation}
{\bf Q}={\bar d}_j {\bf K}^{-1}(c_j^x,c_j^y,1)^T\;.
\end{equation}
By making use of the plane constraint, we can estimate the last parameter $\bar{n}_j^d$ as $\bar{n}_j^d = -\bar{{\bf n}}_j^T {\bf Q}$. 

Finally, let $\tilde{\bf{n}}_j={\bar{\bf n}}_j/\bar{n}_j^d$. 
Given the relative rotation matrix ${\bf R}$ and translation vector ${\bf t}$ between the input and novel views, the homography for region $j$ can be expressed as
\begin{equation}
{\bf H}_j = {\bf K}({\bf R} - {\bf t}\tilde{\bf{n}}_j^T){\bf K}^{-1}.\nonumber
\end{equation}
This lets us compute a homography for every seed region.

\vspace{-0.4cm}
\paragraph{Inverse Image Warping.}
Each resulting homography can be applied to the pixels of the input (source) image. For each source pixel ${\bf x}^s$, this can be written as
\begin{equation}
\lambda \tilde{{\bf x}}_j^t = {\bf H}_j \tilde{{\bf x}}^s\;,
\end{equation}
with $\tilde{{\bf x}}^s$ the pixel location in homogeneous coordinates, and $\lambda$ the corresponding scalar.
While the result of this operation will indeed correspond to a location in the target image (ignoring the fact that some will lie outside the image range), these locations will not correspond to exact, integer pixel coordinates. In our context of generating a novel view, this would significantly complicate the task of obtaining the intensity value at each target pixel, which would require combining the intensities of nearby transformed locations, whose number would vary for each target pixel. 

To address this, instead of following a forward warping strategy (from source image to target image), we rely on an inverse warping (from target image to source image). Specifically, for every target pixel location ${\bf x}_i^t$, we obtain the corresponding source location by relying on the inverse homography ${\bf H}^{-1}_j$ as $\tilde{{\bf x}}_{i,j}^s \propto {\bf H}^{-1}_j \tilde{{\bf x}}_i^t$. We then compute the target intensity value at pixel ${\bf x}_i^t$ by bilinear interpolation as
\begin{equation}
\hat{I}_j^t({\bf x}_i^t) = \sum_{q\in o^i_j}I^s(1-|x_{i,j}^s - x^s_{q,j}|,1-|y_{i,j}^s-y^s_{q,j}|)\;,
\label{eq:biInterpo}
\end{equation}
where $I^s$ is the input source image, and $o_j^i$ denotes the 4-pixel neighborhood of ${\bf x}_{i,j}^s$, which itself is predicted by the inverse homography.

\begin{figure}[t!]
	\centering
	\includegraphics[width=0.85\linewidth]{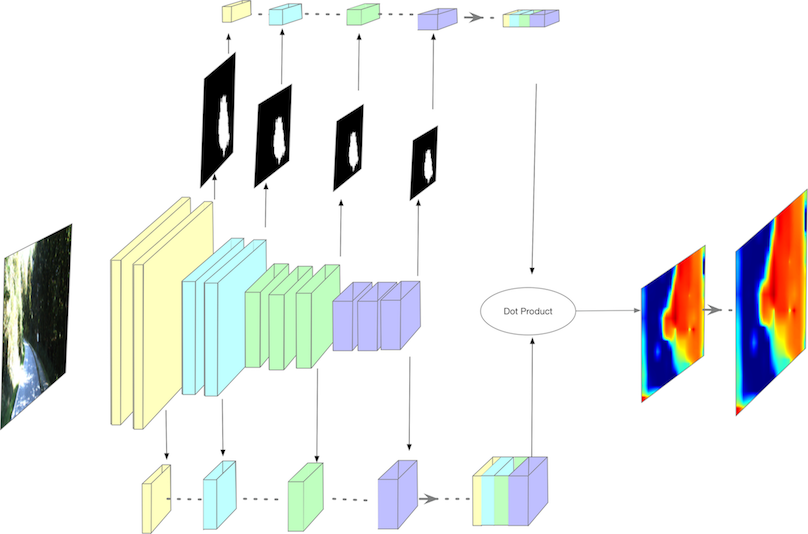} 
	\caption{{\bf Selection Network.} Instead of using hard segmentation masks to combine the candidate synthesized images, we train a network to generate a set of soft selection masks. The network structure follows that of the first 4 blocks of VGG16. We max-pool the corresponding 4 feature maps according to the seed masks and concatenate the resulting 4 feature vectors in a hypercolumn feature.
We then convolve this hypercolumn feature with the concatenated complete feature maps at low resolution, which yields one global heatmap that we upsample to the original image size.
Note that we normalize the pooled features and complete feature maps along the feature dimension. 
	}
\label{fig:selNet}
\vspace{-0.2cm}
\end{figure}
\vspace{-0.4cm}
\paragraph{Selection Network.}

As discussed below, we generate the novel view by assembling the $m$ candidate target images obtained as described above. To this end, we develop a~\emph{selection network} to predict $m$ planar region masks from the input image and seed region masks (Section~\ref{sec:seg}). 
More precisely, for each seed region, we aim to predict a soft selection map indicating the likelihood for every input pixel to be associated to the corresponding homography.
	
Specifically, the structure of our selection network follows that of the first 4 convolutional blocks of VGG16~\cite{simonyan2014very}. As shown in Fig~\ref{fig:selNet}, each seed region mask is used to max-pool the corresponding 4 feature maps. We then concatenate the resulting 4 features to form a hypercolumn~\cite{hariharan2015hypercolumns} feature, which we convolve with the concatenated complete feature maps at the lower resolution. This yields a low-resolution heat map, which we upsample to the original image size. The resulting heat map indicates a notion of similarity between the features at every pixel and the one pooled over the seed region. This procedure is performed individually for the $m$ seed regions, but using shared network parameters.
Note that the resulting $m$ selection maps are defined in the input view, and we thus apply our inverse warping procedure to compute them in the novel view.
\vspace{-0.4cm}
\paragraph{Novel View Prediction.}
Given the selection maps $\{\tilde{M}_j\}$, we first compute a normalized transformed mask for the novel view as
\begin{footnotesize}
	\begin{equation}
	\hat{M}_j^t({\bf x}_i^t) = \frac{\sum\limits_{q\in o^i_j}\tilde{M}_j(1-|x^s_{i,j} - x^s_{q,j}|,1-|y^s_{i,j}-y^s_{q,j}|)+\epsilon}{\sum\limits_{k=1}^m\sum\limits_{q\in o^i_k}(\tilde{M}_k(1-|x^s_{i,k} - x^s_{q,k}|)(1-|y^s{i,k}-y^s_{q,k}|)+\epsilon)}\;.
	\label{eq:mask_trans}
	\end{equation}
\end{footnotesize}
Note that the resulting transformed masks are not binary, but rather provide weights to combine the $m$ estimated target images. To account for the fact that some pixels will be warped outside the input image with all $m$ homographies, we make use of a small constant $\epsilon$, which prevents division by 0 in the normalization process and yields uniform weights for such pixels. In our experiments, we set $\epsilon=0.0001$.
We compute the novel view as
\begin{equation}
\hat{I}^t({\bf x}_i^t) = \sum_{j=1}^m\hat{I}_j^t({\bf x}_i^t)\cdot \hat{M}_j^t({\bf x}_i^t)\;.
\label{eq:pred}
\end{equation}
Note that some of the pixels in the output view will be mapped outside the input image by all homographies. In the simplest version of our approach, we fill in the intensity of each such pixel by using the value at the nearest pixel in the input image. In Section~\ref{sec:refine}, we introduce a refinement network that produces more realistic predictions.

\vspace{-0.3cm}
\subsubsection{Learning}
\vspace{-0.2cm}


The novel view predicted using Eq.~\ref{eq:pred} is a function of the homographies, which themselves are functions of the normal and depth estimates, and of the selection masks, which in turn depend on the depth and normal branch parameters ${\bf W}_{d},{\bf W}_{n}$, and selection network  parameters ${\bf W}_{s}$, respectively. Altogether, the prediction can then be thought of as a function of the parameters ${\bf W}=\{{\bf W}_d,{\bf W}_n, {\bf W}_s\}$ given an input image $I^s$, and a relative pose ${\bf P}$, encompassing the 3D rotation, translation and camera intrinsics, and the segmentation seed region masks $M$.

All the operations described above are differentiable. The least obvious cases are the bilinear interpolations of Eqs.~\ref{eq:biInterpo} and~\ref{eq:mask_trans}, and the use of the inverse homography. For the former ones, we refer the reader to~\cite{jaderberg2015spatial}, who showed that the (sub)-gradient of bilinear interpolation with respect to ${\bf W}$, could be efficiently computed. For the latter case, we propose to exploit the Sherman-Morrison formula~\cite{press2007numerical}, provided in the supplementary material, to avoid having to explicitly compute the inverse of the homography.

%

In our context, this formula lets us express the inverse of the homography analytically as follows. Let
\begin{eqnarray}
{\tilde {\bf H}}^{-1} = {\bf R}^\mathrm{T} +\frac{{\bf R}^\mathrm{T}{\bf t}{\tilde {\bf n}}^\mathrm{T}{\bf R}^\mathrm{T}}{1-{\tilde {\bf n}}^\mathrm{T}{\bf R}^\mathrm{T}{\bf t}}\;.
\end{eqnarray}
Then, we have ${\bf H}^{-1} = {\bf K}{\tilde {\bf H}}^{-1}{\bf K}^{-1}$.
This formulation makes it easy to compute the gradient of the inverse homography w.r.t. the estimated depth and normals, and thus to train our model using backpropagation.

To this end, we make use of an $\ell_1$ loss between the true target image and the estimated one. Given $N$ training samples, learning can then be expressed as
\begin{equation}
\min_{{\bf W}}\frac{1}{N}\sum_{i=1}^N\|I^t_i - \hat{I}^t_i(I^s_i,P_i,M_i,{\bf W})\|_1 \;,
\end{equation}
where $I^t_i$ is the ground-truth novel view, and where, with a slight abuse of notation, we denote the segmentation mask for sample $i$ as $M_i$. More details about optimization are provided in Section~\ref{sec:exp}.


\vspace{-0.2cm}
\subsubsection{Obtaining Seed Regions}
\label{sec:seg}
\vspace{-0.2cm}
Throughout  our framework, we assume to be given $m$ segmentation masks as input, corresponding to the $m$ planes we use to represent the scene. To extract these masks, we make use of the following simple, yet effective strategy. We first over-segment the image into superpixels using SLIC~\cite{achanta2012slic}. For each superpixel, we then extract its RGB value and center location as features and use~\emph{K-means}
to cluster the superpixels into $m$ regions. This strategy has the advantage over learning-based segmentation masks of generating compact regions, which are better suited to estimating the corresponding plane parameters. Furthermore, as evidenced by our experiments, it allows us to obtain accurate synthesized views that respect the scenes 3D structure.

\subsection{Refinement Network}
\label{sec:refine}
\vspace{-0.1cm}

\begin{figure}[t!]
	\vspace{-0.1cm}
	\includegraphics[width=0.95\linewidth]{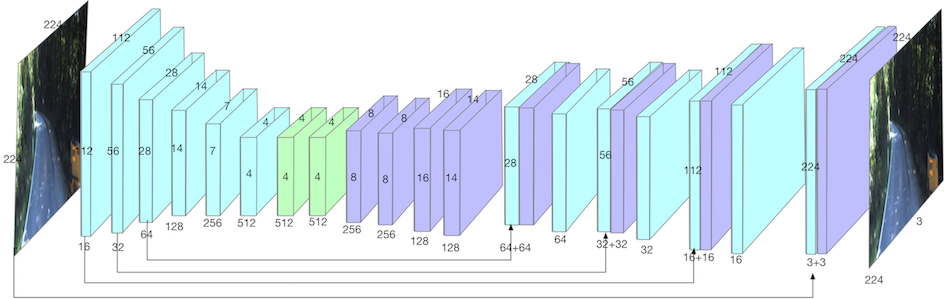}
	\caption{{\bf Refinement Network.} Our refinement network adopts an encoder-decoder structure with skip connections.  The blue blocks denote convolutions with stride two followed by batch normalization and leaky ReLU. The green blocks denote convolutions with stride one followed by batch normalization and leaky ReLU. The purple blocks denote deconvolutions followed by batch normalization and ReLU.}
	\label{fig:RefineNet}
	\vspace{-0.3cm}
\end{figure}

Our region-aware geometric-transformation network produces a novel view image that preserves the local geometric structures of the scene. While geometric transformations can synthesize regions that appear both in the input and novel views, it cannot handle the regions that are only present in the novel view, i.e., that were hidden in the input view. To address this, inspired by~\cite{park2017transformation}, we make use of the encoder-decoder refinement network depicted by Fig.~\ref{fig:RefineNet}. While the structure of this network is the same as in~\cite{park2017transformation}, we make use of a different, simpler loss function to train it.

Specifically, let $L_p$ denote the mean pixel $\ell_1$ error. We then define the loss of our refinement network as
\vspace{-0.2cm}
\begin{equation}
L_t = L_p+\lambda L_f\;,
\label{eq:refine}
\end{equation}
where $L_f$ is a feature $\ell_1$ loss. That is, it corresponds to an $\ell_1$ loss between features extracted from a fixed VGG-19 network, pre-trained for classification on ImageNet. In particular, we concatenate features from the `conv1$\_$2', `conv2$\_$2', `conv3$\_$2', `conv4$\_$2' and `conv5$\_$2' layers of VGG-19. This strategy has proven effective in~\cite{chen2017photographic} in the context of image-to-image translation. In particular, it has the advantage over~\cite{park2017transformation} of not relying on a generative adversarial network, which are known to be hard to train. As shown in our results, this refinement network not only hallucinates the missing parts of the synthesized images, but it also removes the blur arising from combining multiple warped images.

\vspace{-0.2cm} 
\section{Experiments}
\label{sec:exp}
\vspace{-0.2cm} 


We evaluate our approach both quantitatively and qualitatively on the challenging urban KITTI odometry dataset~\cite{Geiger2012CVPR}, which depicts complex scenes with rich structure and dynamic objects, and on the large indoor scene ScanNet dataset~\cite{dai2017scannet}, which covers diverse scene types. We compare our approach with the state-of-the-art single-image view synthesis algorithm of~\cite{zhou2016view} for real-world scenes\footnote{Note that, as discussed in Section~\ref{sec:related}, the transformation-grounded network of~\cite{park2017transformation} focuses on single-object novel view synthesis.}. Furthermore, we also report the results of a depth-based baseline consisting of using the predictions of our depth stream warped to the new pose, followed by bicubic interpolation to obtain a complete image. 

\begin{figure}[t!]
	\vspace{-0.0cm}
	\includegraphics[width=0.90\linewidth]{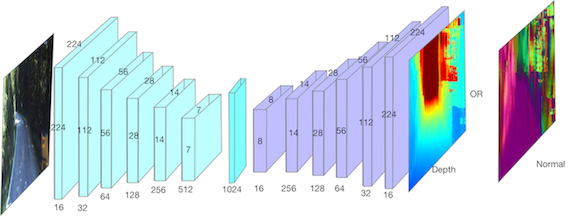}
	\caption{{\bf Encoder-decoder network for depth or normal prediction on KITTI.} Both our depth and normal streams make use of this architecture. However, they rely on different parameters. 
	}
	\label{fig:ed}
	\vspace{-0.3cm}
\end{figure}

\vspace{-0.1cm}
\subsection{Experimental Setup} 
\vspace{-0.1cm} 
\noindent{\bf KITTI Dataset.}~For the comparison with~\cite{zhou2016view} to be fair, we adopt the same data splits as them. Namely, we use the video sequences with index 0 to 8 as training set, and 9 to 10 as test set. We then generate our training and test pair in the following way, similar to that of~\cite{zhou2016view}: For each image in a sequence, we randomly sample a frame number for the input image and for the target image such that they are separated by at most $\pm 10$ frames. 

\noindent{\bf ScanNet Dataset.}~We make use of the training, validation and test splits provided with ScanNet. In particular, we use 405 training sequences to learn our model and 312 sequences from the test set for testing.  We form the input-target pairs in the same manner as for KITTI.
In total, we use 30000 training pairs and 5000 test pairs.

We resize the images from both datasets to $224\times 224 \times 3.$ to match that of~\cite{zhou2016view}. To obtain the segmentation masks, we first oversegment each image into $400$ SLIC~\cite{achanta2012slic} superpixels and cluster them into $m = 16$ regions, as described in Section~\ref{sec:seg}. This represents a good trade-off between the accuracy of our piece-wise planar representation on the training data and the memory consumption of our method.
In practice, this proved sufficient to yield realistic novel views. 

\begin{figure*}[t!]
	\vspace{-0.2cm}
	\begin{small}
	\begin{center}
		\begin{tabular}{ccccc}
			\hspace{-0.0cm}Input view &
			\hspace{-0.05cm} App. Flow~\cite{zhou2016view} &
			\hspace{-0.05cm} Ours-Geo &
			\hspace{-0.05cm} Ours-Full &
			\hspace{-0.05cm} Ground-truth \\
			\hspace{-0.0cm}\includegraphics[width=0.16\linewidth]{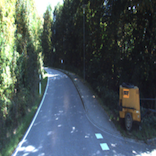}&
			 \hspace{-0.05cm}\includegraphics[width=0.16\linewidth]{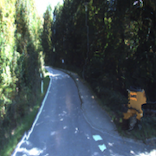}&
			\hspace{-0.05cm}\includegraphics[width=0.16\linewidth]{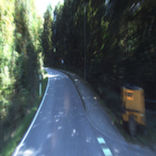}&
			\hspace{-0.05cm}\includegraphics[width=0.16\linewidth]{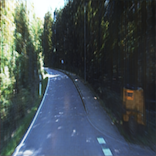}&
			\hspace{-0.05cm}\includegraphics[width=0.16\linewidth]{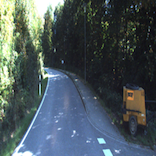}\\
			\hspace{-0.0cm}\includegraphics[width=0.16\linewidth]{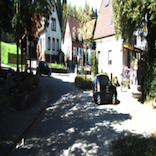}&
			\hspace{-0.05cm}\includegraphics[width=0.16\linewidth]{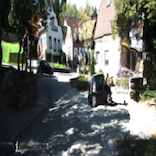}&
			\hspace{-0.05cm}\includegraphics[width=0.16\linewidth]{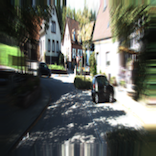}&
			\hspace{-0.05cm}\includegraphics[width=0.16\linewidth]{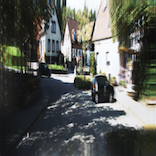}&
			\hspace{-0.05cm}\includegraphics[width=0.16\linewidth]{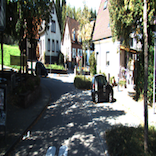}\\
			\hspace{-0.0cm}\includegraphics[width=0.16\linewidth]{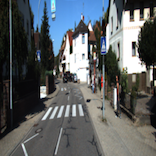}&
			\hspace{-0.05cm}\includegraphics[width=0.16\linewidth]{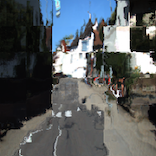}&
			\hspace{-0.05cm}\includegraphics[width=0.16\linewidth]{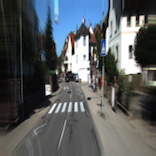}&
			\hspace{-0.05cm}\includegraphics[width=0.16\linewidth]{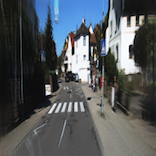}&			
			\hspace{-0.05cm}\includegraphics[width=0.16\linewidth]{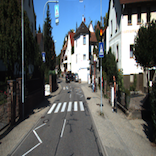}
		\end{tabular}
		\end{center}
	\end{small}
	\vspace{-0.35cm}
	\caption{{\bf Qualitative comparison of our approach with the appearance flow method of~\cite{zhou2016view} on KITTI.} While appearance flow yields artifacts, our approach, which reasons about 3D geometry, yields more realistic results. This is noticeable, for instance, by looking at the bottom right part of the first image and at the buildings in the other images. }
	\label{fig:kittires}
	\vspace{-0.2cm}
\end{figure*}

\vspace{-0.1cm}
\subsection{Training Procedure}
We train our model in a stage-wise manner: First, the depth and normal branches, then the selection network given fixed depth and normal branches, and finally the refinement network while rest of the framework is fixed. We tried to then fine-tune the entire network end-to-end, but did not observe any significant improvement.

\vspace{-0.4cm}
\paragraph{Training the depth and normal networks.}
For the indoor ScanNet dataset, we were able to directly use the network of~\cite{eigen2015predicting}, which predicts both depth and normals. This network was pre-trained on NYU-v2~\cite{Silberman:ECCV12}, and we simply fine-tuned it on our data. In particular, since ScanNet does not provide ground-truth normals, we fit a plane to each SLIC superpixel, and assigned the corresponding normal to all its pixels. The fine-tuned network yields a relative depth error of 0.236. We do not report the normal error, since the ground-truth normals were obtained from the depth maps.

For KITTI, we were unfortunately unable to train an equivalent model from scratch. Therefore, we relied on the simpler encoder-decoder network of Fig.~\ref{fig:ed}, which is more compact and easier to train. To this end, we made use of the $\ell_1$ loss for the inverse depth and of the negative inner product as a normal loss. Note that KITTI only provides sparse ground-truth depth maps. While this is sufficient to train the depth branch, it does not allow us to generate ground-truth normals as in ScanNet. To this end, we used the stereo framework of~\cite{zbontar2015computing} to generate dense depth maps, which we used, in turn, to obtain normal maps using  superpixels. The final depth network yields a relative error of 0.274.

Note that we analyze the influence of the depth and normal prediction accuracy on our final novel view synthesis results in our results section.

\begin{table}[t]
	\centering
	\begin{tabular}{c | c |c}
		\hline
		Method & $\ell_1$-KITTI&$\ell_1$-ScanNet\\ 
		\hline
		App. flow~\cite{zhou2016view} & 0.471&{-}\\
		Depth-branch & 0.668&0.217\\
		Ours-Geo & {\bf 0.340}&{\bf 0.167}\\
		Ours-Full & 0.345&0.176\\
		\hline
	\end{tabular}
	\vspace{0.1cm}
	\caption{{\bf Quantitative evaluation on KITTI and ScanNet.} We compare our approach with the state-of-the-art method of~\cite{zhou2016view} and our baseline based on our depth estimates. Our approach significantly outperforms the baselines, thus achieving state-of-the-art performance on these datasets.}
	\label{Tab:oursCompKitti}
	\vspace{-0.3cm}
\end{table}

\vspace{-0.3cm}
\paragraph{Training the selection network.}
The selection network takes the predicted depth and normals, together with the image, relative pose and seed regions, as input to synthesize the novel view. Since we do not have ground-truth labels for the selection maps, we therefore directly trained the selection network using the mean pixel $\ell_1$ error as a loss. 

\vspace{-0.3cm}
\paragraph{Training the refinement network.}
The refinement network aims to improve an initial synthesized view.
We train it using the loss of Eq.~\ref{eq:refine}, with $\lambda = 0.01$. 

We implemented our model in tensorflow and trained it on two NVIDIA Tesla P100, each with 16GB memory.  We used mini-batches of size 10, and employed the ADAM solver with a learning rate of $0.0001$,and the default values $\beta_1=0.9$ and $\beta_2=0.999$. 
We will make our code publicly available upon acceptance of the paper.

\vspace{-0.1cm}
\subsection{Results}
\vspace{-0.15cm}
In Table~\ref{Tab:oursCompKitti}, we compare our  approach, both without (Ours-Geo) and with (Ours-Full) refinement network, with the state-of-the-art appearance flow technique of ~\cite{zhou2016view} on KITTI and  ScanNet, based on the mean pixel $\ell_1$ error metric.  Note that our approach outperforms the baseline that uses our depth estimates, without explicitly modeling the scene structure, by a large margin. This evidences the importance of accounting for 3D scene structure. Our approach also significantly outperforms the state-of-the-art appearance flow method on KITTI.\footnote{Note that, because the training code for appearance flow is not available, we had to re-implement it, and despite confirming that our implementation was correct using the KITTI dataset, we were unable to make training converge on ScanNet.} This again shows the benefits of modeling geometry, as done by our region-aware geometric-transform network. Interestingly, the refinement network tends to slightly degrade the novel view accuracy.

 \begin{figure}[t!]
	\vspace{-0.3cm}
	\begin{small}
	\begin{center}
		\begin{tabular}{cccc}
			\hspace{-0.3cm}\begin{sideways}\hspace{0.55cm}Input view \end{sideways} &
			\hspace{-0.2cm}\includegraphics[width=0.27\linewidth]{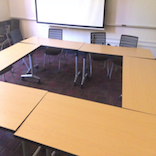}&
			\hspace{-0.2cm}\includegraphics[width=0.27\linewidth]{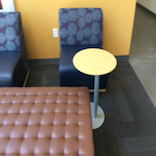}&
			\hspace{-0.2cm}\includegraphics[width=0.27\linewidth]{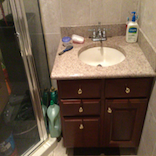}\\
			\hspace{-0.3cm}\begin{sideways}\hspace{0.55cm}Ours-Geo \end{sideways}&
			\hspace{-0.2cm}\includegraphics[width=0.27\linewidth]{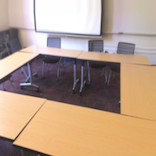}&
			\hspace{-0.2cm}\includegraphics[width=0.27\linewidth]{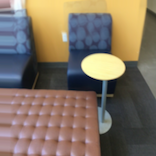}&
			\hspace{-0.2cm}\includegraphics[width=0.27\linewidth]{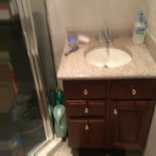}\\
			\hspace{-0.3cm}\begin{sideways}\hspace{0.55cm}Ours-Full \end{sideways}&
			\hspace{-0.2cm}\includegraphics[width=0.27\linewidth]{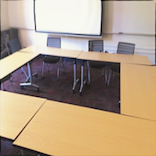}&
			\hspace{-0.2cm}\includegraphics[width=0.27\linewidth]{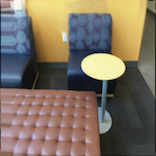}&
			\hspace{-0.2cm}\includegraphics[width=0.27\linewidth]{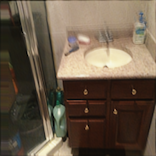}\\			
			\hspace{-0.3cm}\begin{sideways}\hspace{0.5cm}Ground-truth \end{sideways} &
			\hspace{-0.2cm}\includegraphics[width=0.27\linewidth]{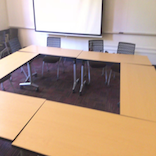}&
			\hspace{-0.2cm}\includegraphics[width=0.27\linewidth]{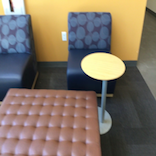}&
			\hspace{-0.2cm}\includegraphics[width=0.27\linewidth]{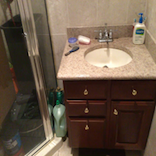}
		\end{tabular}
		\end{center}
	\end{small}
	\vspace{-0.37cm}
	\caption{{\bf Qualitative results of our approach on ScanNet.} }
	\label{fig:scannetres}
	\vspace{-0.0cm}
\end{figure}
\begin{table}[t]
	\centering
	\small
	\begin{tabular}{c|c|c|c|c|c|c}
		\hline
		gtDep&gtNor&estDep&estNor&Seed&SelMap&$\ell_1$\\
		\hline		
		\cmark&\cmark&\xmark&\xmark&\cmark&\xmark&0.357\\
		\hline
		\cmark&\cmark&\xmark&\xmark&\xmark&\cmark&0.329\\
		\hline
		\xmark&\xmark&\cmark&\cmark&\cmark&\xmark&0.373\\
		\hline
		\xmark&\xmark&\cmark&\cmark&\xmark&\cmark&0.340\\
		\hline
	\end{tabular}
	\vspace{0.0cm}
\caption{{\bf Influence of the quality of the depth and normal estimates and of learning the selection maps on KITTI.} From left to right: gtDep and gtNor denote the ground-truth depth and normals, respectively; estDep and estNor denote the estimated depth and normals, respectively; Seed and SelMap denote the hard-segmentations corresponding to the seed region and the selection maps obtained with our selection network, respectively.}
\label{tab:comparisonKitti}
\vspace{-0.4cm}
\end{table}

However, when looking at the qualitative comparison in Figs.~\ref{fig:intro},~\ref{fig:kittires} and~\ref{fig:scannetres}, we can see that our complete model (Ours-Full) yields more realistic novel views than both Ours-Geo and appearance flow~\cite{zhou2016view}. Note that, by not leveraging structure, appearance flow yields to unrealistic artifacts. By contrast, the results of our approach that exploits 3D geometry look more natural. This, for instance, can be observed by looking at the bottom-right corner of the first image in Fig.~\ref{fig:kittires}, where we better model the shape of the object,  and at the buildings in the other images.

In Table~\ref{tab:comparisonKitti},
we analyze the influence of the quality of the depth and normal estimates, and the effect of learning the selection maps. In particular, we compare the error obtained when using the ground-truth depth and normals instead of the predicted ones, and when using the seed regions as 'hard' segmentation masks instead of the learnt selection maps. In both cases, the best results are obtained by using the ground-truth depth and normals in conjunction with our selection maps, followed by using the estimated depth and normals with our selection maps. This shows (i) the importance of learning the combination of the multiple synthesized candidates; and (ii) that the results of our approach will further improve as progress in single-image depth and normal prediction is made. A similar table for ScanNet is provided in the supplementary material.

In Fig.~\ref{fig:selectionMapimgs}, we illustrate what the selection network learns. To this end, we show the initial seed region overlaid with input image, and the likelihood of the pixels to be associated to this plane, predicted by the selection network. From the examples, we can see that the selection network extends the initial seed regions to larger planes of semantically and visually coherent pixels, such as a larger tree regions.
\begin{figure}[t!]
	\vspace{-0.5cm}
	\begin{center}
	\begin{tabular}{ccc}
		\hspace{0cm} & \hspace{-0.0cm}Input image &\\
		&\hspace{-0.0cm}\includegraphics[width=0.27\linewidth,height=0.22\linewidth]{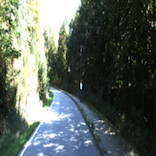}&\vspace{0.05cm}\\
		\hspace{-0.0cm}Seed Region& 	\hspace{-0.2cm}Selection Map &	\hspace{-0.2cm}  Overlay Image \\
		\hspace{-0.0cm}\includegraphics[width=0.27\linewidth,height=0.22\linewidth]{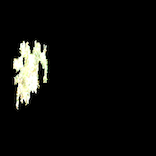}& \hspace{-0.3cm}\includegraphics[width=0.27\linewidth,height=0.22\linewidth]{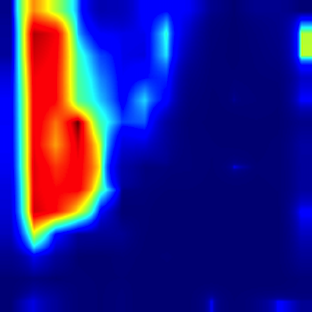}&
		\hspace{-0.3cm}\includegraphics[width=0.27\linewidth,height=0.22\linewidth]{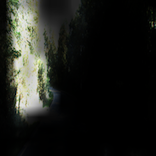}\\
		\hspace{-0.0cm}\includegraphics[width=0.27\linewidth,height=0.22\linewidth]{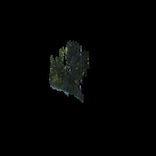}& \hspace{-0.3cm}\includegraphics[width=0.27\linewidth,height=0.22\linewidth]{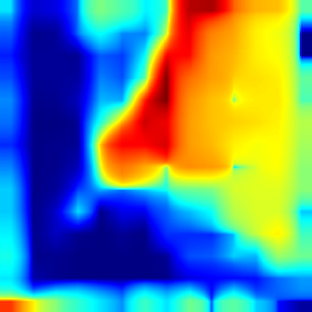}&
		\hspace{-0.3cm}\includegraphics[width=0.27\linewidth,height=0.22\linewidth]{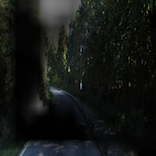}\\
		\hspace{-0.05cm} \includegraphics[width=0.27\linewidth,height=0.22\linewidth]{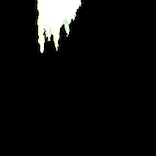}& \hspace{-0.3cm}\includegraphics[width=0.27\linewidth,height=0.22\linewidth]{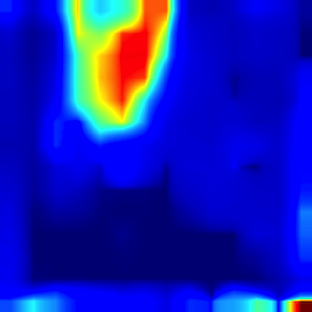}&
		\hspace{-0.3cm}\includegraphics[width=0.27\linewidth,height=0.22\linewidth]{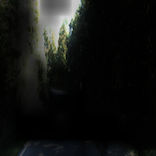}
	\end{tabular}
	\end{center}
	\vspace{-0.3cm}
	\caption{{\bf Sample seed regions and predicted selection maps in the input view.} From left to right: the seed region, predicted selection map and selection map overlaid on the input image, showing that the corresponding region is close to planar. Red indicates a high likelihood for a pixel to belong to the plane defined by the seed region and blue to a low likelihood.}
	\label{fig:selectionMapimgs}
	\vspace{-0.32cm}
\end{figure}

\vspace{-0.4cm}
\section{Conclusion}
\vspace{-0.3cm}
We have introduced a geometry-aware deep learning framework for novel view synthesis from a single image. Our approach models the scene with a fixed number of planes, and learns to predict homographies, which, in conjunction with a predicted selection map and a desired relative pose, let us generate the novel view. Our experiments on the challenging KIITI and ScanNet datasets
have demonstrated the benefits of our approach; by leveraging 3D geometry, our method yields predictions that better match the scene structure, and thus outperforms the state-of-the-art single-image novel view synthesis techniques.
Training the depth branch of our framework currently relies on ground-truth depth maps. In the future, we will investigate the use of weakly-supervised depth prediction methods~\cite{garg2016unsupervised,godard2016unsupervised,zhou2017unsupervised} that only exploit two views
to perform this task.


\vspace{-0.4mm}
\noindent{\bf Acknowledgments} 
This work was done when the first author was working in Data61, CSIRO, Australia.The Titan X used for this research was donated by the NVIDIA Corporation. 

{\small
\bibliographystyle{ieee}
\bibliography{reference}

\begin{thebibliography}{10}\itemsep=-1pt

\bibitem{achanta2012slic}
R.~Achanta, A.~Shaji, K.~Smith, A.~Lucchi, P.~Fua, and S.~S{\"u}sstrunk.
\newblock Slic superpixels compared to state-of-the-art superpixel methods.
\newblock {\em IEEE transactions on pattern analysis and machine intelligence},
  34(11):2274--2282, 2012.

\bibitem{chaurasia2013depth}
G.~Chaurasia, S.~Duchene, O.~Sorkine-Hornung, and G.~Drettakis.
\newblock Depth synthesis and local warps for plausible image-based navigation.
\newblock {\em ACM Transactions on Graphics (TOG)}, 32(3):30, 2013.

\bibitem{chen2017photographic}
Q.~Chen and V.~Koltun.
\newblock Photographic image synthesis with cascaded refinement networks.
\newblock In {\em ICCV}, 2017.

\bibitem{dai2017scannet}
A.~Dai, A.~X. Chang, M.~Savva, M.~Halber, T.~Funkhouser, and M.~Nie{\ss}ner.
\newblock Scannet: Richly-annotated 3d reconstructions of indoor scenes.
\newblock In {\em Proc. Computer Vision and Pattern Recognition (CVPR), IEEE},
  2017.

\bibitem{eigen2015predicting}
D.~Eigen and R.~Fergus.
\newblock Predicting depth, surface normals and semantic labels with a common
  multi-scale convolutional architecture.
\newblock In {\em Proceedings of the IEEE International Conference on Computer
  Vision}, pages 2650--2658, 2015.

\bibitem{flynn2016deepstereo}
J.~Flynn, I.~Neulander, J.~Philbin, and N.~Snavely.
\newblock Deepstereo: Learning to predict new views from the world's imagery.
\newblock In {\em Proceedings of the IEEE Conference on Computer Vision and
  Pattern Recognition}, pages 5515--5524, 2016.

\bibitem{furukawa2010accurate}
Y.~Furukawa and J.~Ponce.
\newblock Accurate, dense, and robust multiview stereopsis.
\newblock {\em IEEE transactions on pattern analysis and machine intelligence},
  32(8):1362--1376, 2010.

\bibitem{garg2016unsupervised}
R.~Garg, G.~Carneiro, and I.~Reid.
\newblock Unsupervised cnn for single view depth estimation: Geometry to the
  rescue.
\newblock In {\em European Conference on Computer Vision}, pages 740--756.
  Springer, 2016.

\bibitem{Geiger2012CVPR}
A.~Geiger, P.~Lenz, and R.~Urtasun.
\newblock Are we ready for autonomous driving? the kitti vision benchmark
  suite.
\newblock In {\em Conference on Computer Vision and Pattern Recognition
  (CVPR)}, 2012.

\bibitem{godard2016unsupervised}
C.~Godard, O.~Mac~Aodha, and G.~J. Brostow.
\newblock Unsupervised monocular depth estimation with left-right consistency.
\newblock {\em arXiv preprint arXiv:1609.03677}, 2016.

\bibitem{hariharan2015hypercolumns}
B.~Hariharan, P.~Arbel{\'a}ez, R.~Girshick, and J.~Malik.
\newblock Hypercolumns for object segmentation and fine-grained localization.
\newblock In {\em Proceedings of the IEEE Conference on Computer Vision and
  Pattern Recognition}, pages 447--456, 2015.

\bibitem{hoiem2005automatic}
D.~Hoiem, A.~A. Efros, and M.~Hebert.
\newblock Automatic photo pop-up.
\newblock {\em ACM transactions on graphics (TOG)}, 24(3):577--584, 2005.

\bibitem{horry1997tour}
Y.~Horry, K.-I. Anjyo, and K.~Arai.
\newblock Tour into the picture: using a spidery mesh interface to make
  animation from a single image.
\newblock In {\em Proceedings of the 24th annual conference on Computer
  graphics and interactive techniques}, pages 225--232. ACM
  Press/Addison-Wesley Publishing Co., 1997.

\bibitem{jaderberg2015spatial}
M.~Jaderberg, K.~Simonyan, A.~Zisserman, and K.~Kavukcuoglu.
\newblock Spatial transformer networks.
\newblock In {\em Advances in Neural Information Processing Systems}, pages
  2017--2025, 2015.

\bibitem{ji2017deep}
D.~Ji, J.~Kwon, M.~McFarland, and S.~Savarese.
\newblock Deep view morphing.
\newblock In {\em Proc. Computer Vision and Pattern Recognition (CVPR), IEEE},
  2017.

\bibitem{kulkarni2015deep}
T.~D. Kulkarni, W.~F. Whitney, P.~Kohli, and J.~Tenenbaum.
\newblock Deep convolutional inverse graphics network.
\newblock In {\em Advances in Neural Information Processing Systems}, pages
  2539--2547, 2015.

\bibitem{liu2009content}
F.~Liu, M.~Gleicher, H.~Jin, and A.~Agarwala.
\newblock Content-preserving warps for 3d video stabilization.
\newblock {\em ACM Transactions on Graphics (TOG)}, 28(3):44, 2009.

\bibitem{mcmillan1995plenoptic}
L.~McMillan and G.~Bishop.
\newblock Plenoptic modeling: An image-based rendering system.
\newblock In {\em Proceedings of the 22nd annual conference on Computer
  graphics and interactive techniques}, pages 39--46. ACM, 1995.

\bibitem{Silberman:ECCV12}
P.~K. Nathan~Silberman, Derek~Hoiem and R.~Fergus.
\newblock Indoor segmentation and support inference from rgbd images.
\newblock In {\em ECCV}, 2012.

\bibitem{park2017transformation}
E.~Park, J.~Yang, E.~Yumer, D.~Ceylan, and A.~C. Berg.
\newblock Transformation-grounded image generation network for novel 3d view
  synthesis.
\newblock In {\em CVPR}, 2017.

\bibitem{press2007numerical}
W.~H. Press.
\newblock {\em Numerical recipes 3rd edition: The art of scientific computing}.
\newblock Cambridge university press, 2007.

\bibitem{rematas2016novel}
K.~Rematas, C.~Nguyen, T.~Ritschel, M.~Fritz, and T.~Tuytelaars.
\newblock Novel views of objects from a single image.
\newblock {\em arXiv preprint arXiv:1602.00328}, 2016.

\bibitem{simonyan2014very}
K.~Simonyan and A.~Zisserman.
\newblock Very deep convolutional networks for large-scale image recognition.
\newblock {\em arXiv preprint arXiv:1409.1556}, 2014.

\bibitem{tatarchenko2016multi}
M.~Tatarchenko, A.~Dosovitskiy, and T.~Brox.
\newblock Multi-view 3d models from single images with a convolutional network.
\newblock In {\em European Conference on Computer Vision}, pages 322--337.
  Springer, 2016.

\bibitem{woodford2007new}
O.~J. Woodford, I.~D. Reid, P.~H. Torr, and A.~W. Fitzgibbon.
\newblock On new view synthesis using multiview stereo.
\newblock In {\em BMVC}, pages 1--10, 2007.

\bibitem{xie2016deep3d}
J.~Xie, R.~Girshick, and A.~Farhadi.
\newblock Deep3d: Fully automatic 2d-to-3d video conversion with deep
  convolutional neural networks.
\newblock In {\em European Conference on Computer Vision}, pages 842--857.
  Springer, 2016.

\bibitem{zbontar2015computing}
J.~Zbontar and Y.~LeCun.
\newblock Computing the stereo matching cost with a convolutional neural
  network.
\newblock In {\em Proceedings of the IEEE Conference on Computer Vision and
  Pattern Recognition}, pages 1592--1599, 2015.

\bibitem{zhou2017unsupervised}
T.~Zhou, M.~Brown, N.~Snavely, and D.~G. Lowe.
\newblock Unsupervised learning of depth and ego-motion from video.
\newblock {\em arXiv preprint arXiv:1704.07813}, 2017.

\bibitem{zhou2016view}
T.~Zhou, S.~Tulsiani, W.~Sun, J.~Malik, and A.~A. Efros.
\newblock View synthesis by appearance flow.
\newblock In {\em European Conference on Computer Vision}, pages 286--301.
  Springer, 2016.

\bibitem{zhou2013plane}
Z.~Zhou, H.~Jin, and Y.~Ma.
\newblock Plane-based content preserving warps for video stabilization.
\newblock In {\em Proceedings of the IEEE Conference on Computer Vision and
  Pattern Recognition}, pages 2299--2306, 2013.

\end{thebibliography}
}
\clearpage

\newpage
\section{Sherman-Morrison formula}
We first provide more detail for the Sherman-Morrison formula, which allows us to explicitly compute the inverse of homographies. The Sherman-Morrison formula can be stated as follows:
\vspace{-0.2cm}
\begin{theorem}
Assume {\bf A} is invertible, and {\bf u} and {\bf v} are column vectors. Furthermore, assume $1+{\bf v}^\mathrm{T}{\bf A}^{-1}{\bf u}\neq 0$. Given
\begin{eqnarray} 
{\bf B} = {\bf A}+{\bf u}{\bf v}^\mathrm{T}, 
\label{eq:B}
\end{eqnarray}
the inverse ${\bf B}^{-1}$ can be obtained as 
\begin{eqnarray}
{\bf B}^{-1} = {\bf A}^{-1}-\frac{{\bf A}^{-1}{\bf u}{\bf v}^\mathrm{T}{\bf A}^{-1}}{1+{\bf v}^\mathrm{T}{\bf A}^{-1}{\bf u}}.
\end{eqnarray}
\end{theorem}
In our context, the homography is defined as 
\begin{equation}
{\bf H}= {\bf K}({\bf R} - {\bf t}\tilde{\bf{n}}^\mathrm{T}){\bf K}^{-1}.\nonumber
\end{equation}
Let us first ignore ${\bf K}$ and concentrate on the central part
\begin{equation}
{\tilde{\bf H}} = {\bf R} - {\bf t}\tilde{\bf{n}}^\mathrm{T}\;,
\label{eq:homo}
\end{equation}
where ${\bf R}$ is a rotation matrix and is thus invertible, i.e., ${\bf R}^{-1} = {\bf R}^\mathrm{T}$. Therefore, Eq.~\ref{eq:homo} satisfies the conditions of ${\bf B}$ in Eq.~\ref{eq:B}, and the inverse ${\tilde{{\bf H}}}^{-1}$ can be written as
\begin{eqnarray}
{\tilde {\bf H}}^{-1} = {\bf R}^\mathrm{T} +\frac{{\bf R}^\mathrm{T}{\bf t}{\tilde {\bf n}}^\mathrm{T}{\bf R}^\mathrm{T}}{1-{\tilde {\bf n}}^\mathrm{T}{\bf R}^\mathrm{T}{\bf t}}\;.\nonumber
\end{eqnarray}
Re-introducing ${\bf K}$, and following the standard rule for matrix product inversion, lets us write the inverse ${\bf H}^{-1}$ as 
\begin{eqnarray}
{\bf H}^{-1} = {\bf K}{\tilde {\bf H}}^{-1}{\bf K}^{-1}.\nonumber
\end{eqnarray}

\section{Experiments}
In this section, we provide additional results on the two datasets. We further illustrate the $m=16$ synthesized images obtained from the homographies generated by our method, and show additional examples of the selection maps our network predicts. We then provide the visualisation of our estimated depth and normal maps for KITTI dataset and discuss failure cases of our approach.


\subsection{Additional Results}
\begin{figure*}[t!]
	\vspace{-0.2cm}
	\begin{small}
		\begin{tabular}{ccccc}
			\hspace{-0.0cm}Input view &
			\hspace{-0.05cm} App. Flow~\cite{zhou2016view} &
			\hspace{-0.05cm} Ours-Geo &
			\hspace{-0.05cm} Ours-Full &
			\hspace{-0.05cm} Ground-truth \\
			\hspace{-0.0cm}\includegraphics[width=0.18\linewidth]{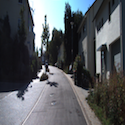}&
			\hspace{-0.05cm}\includegraphics[width=0.18\linewidth]{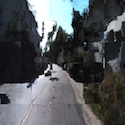}&
			\hspace{-0.05cm}\includegraphics[width=0.18\linewidth]{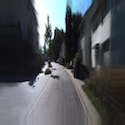}&
			\hspace{-0.05cm}\includegraphics[width=0.18\linewidth]{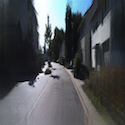}&
			\hspace{-0.05cm}\includegraphics[width=0.18\linewidth]{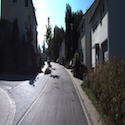}\\
			\hspace{-0.0cm}\includegraphics[width=0.18\linewidth]{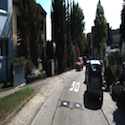}&
			\hspace{-0.05cm}\includegraphics[width=0.18\linewidth]{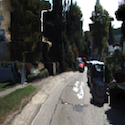}&
			\hspace{-0.05cm}\includegraphics[width=0.18\linewidth]{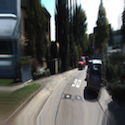}&
			\hspace{-0.05cm}\includegraphics[width=0.18\linewidth]{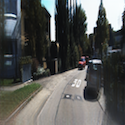}&
			\hspace{-0.05cm}\includegraphics[width=0.18\linewidth]{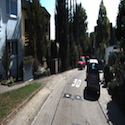}\\
			\hspace{-0.0cm}\includegraphics[width=0.18\linewidth]{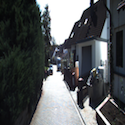}&
			\hspace{-0.05cm}\includegraphics[width=0.18\linewidth]{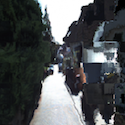}&
			\hspace{-0.05cm}\includegraphics[width=0.18\linewidth]{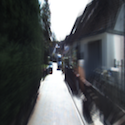}&
			\hspace{-0.05cm}\includegraphics[width=0.18\linewidth]{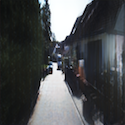}&			
			\hspace{-0.05cm}\includegraphics[width=0.18\linewidth]{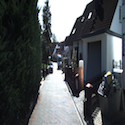}\\
			\hspace{-0.0cm}\includegraphics[width=0.18\linewidth]{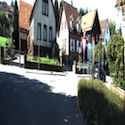}&
			\hspace{-0.05cm}\includegraphics[width=0.18\linewidth]{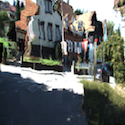}&
			\hspace{-0.05cm}\includegraphics[width=0.18\linewidth]{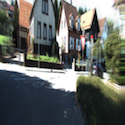}&
			\hspace{-0.05cm}\includegraphics[width=0.18\linewidth]{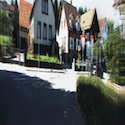}&			
			\hspace{-0.05cm}\includegraphics[width=0.18\linewidth]{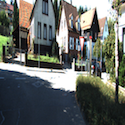}\\
			\hspace{-0.0cm}\includegraphics[width=0.18\linewidth]{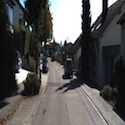}&
			\hspace{-0.05cm}\includegraphics[width=0.18\linewidth]{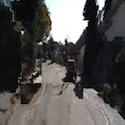}&
			\hspace{-0.05cm}\includegraphics[width=0.18\linewidth]{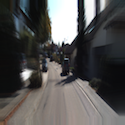}&
			\hspace{-0.05cm}\includegraphics[width=0.18\linewidth]{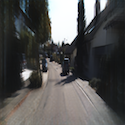}&			
			\hspace{-0.05cm}\includegraphics[width=0.18\linewidth]{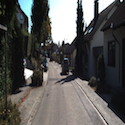}\\
			\hspace{-0.0cm}\includegraphics[width=0.18\linewidth]{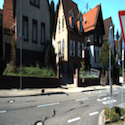}&
			\hspace{-0.05cm}\includegraphics[width=0.18\linewidth]{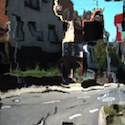}&
			\hspace{-0.05cm}\includegraphics[width=0.18\linewidth]{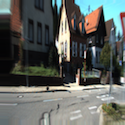}&
			\hspace{-0.05cm}\includegraphics[width=0.18\linewidth]{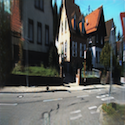}&		
			\hspace{-0.05cm}\includegraphics[width=0.18\linewidth]{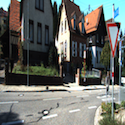}\\
			\hspace{-0.0cm}\includegraphics[width=0.18\linewidth]{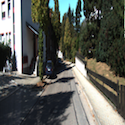}&
			\hspace{-0.05cm}\includegraphics[width=0.18\linewidth]{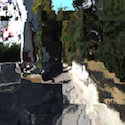}&
			\hspace{-0.05cm}\includegraphics[width=0.18\linewidth]{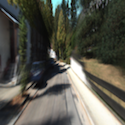}&
			\hspace{-0.05cm}\includegraphics[width=0.18\linewidth]{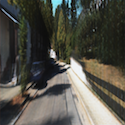}&		
			\hspace{-0.05cm}\includegraphics[width=0.18\linewidth]{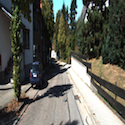}		
		\end{tabular}
	\end{small}
	\caption{{\bf Qualitative comparison of our approach with the appearance flow method of~\cite{zhou2016view} on KITTI.} While appearance flow yields artifacts, our approach, which reasons about 3D geometry, yields more realistic results.}
	\label{fig:kittires_add1}
	\vspace{-0.2cm}
\end{figure*}

\begin{figure*}[t!]
	\vspace{-0.2cm}
	\begin{small}
		\begin{tabular}{ccccc}
			\hspace{-0.0cm}Input view &
			\hspace{-0.05cm} App. Flow~\cite{zhou2016view} &
			\hspace{-0.05cm} Ours-Geo &
			\hspace{-0.05cm} Ours-Full &
			\hspace{-0.05cm} Ground-truth \\
			\hspace{-0.0cm}\includegraphics[width=0.18\linewidth]{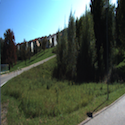}&
			\hspace{-0.05cm}\includegraphics[width=0.18\linewidth]{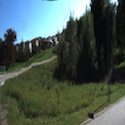}&
			\hspace{-0.05cm}\includegraphics[width=0.18\linewidth]{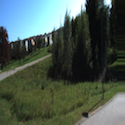}&
			\hspace{-0.05cm}\includegraphics[width=0.18\linewidth]{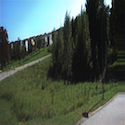}&
			\hspace{-0.05cm}\includegraphics[width=0.18\linewidth]{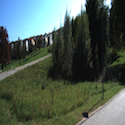}\\
			\hspace{-0.0cm}\includegraphics[width=0.18\linewidth]{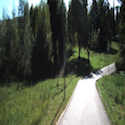}&
			\hspace{-0.05cm}\includegraphics[width=0.18\linewidth]{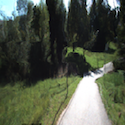}&
			\hspace{-0.05cm}\includegraphics[width=0.18\linewidth]{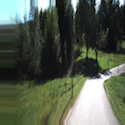}&
			\hspace{-0.05cm}\includegraphics[width=0.18\linewidth]{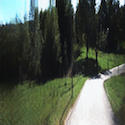}&
			\hspace{-0.05cm}\includegraphics[width=0.18\linewidth]{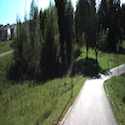}\\
			\hspace{-0.0cm}\includegraphics[width=0.18\linewidth]{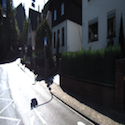}&
			\hspace{-0.05cm}\includegraphics[width=0.18\linewidth]{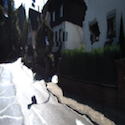}&
			\hspace{-0.05cm}\includegraphics[width=0.18\linewidth]{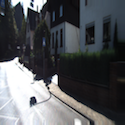}&
			\hspace{-0.05cm}\includegraphics[width=0.18\linewidth]{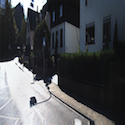}&			
			\hspace{-0.05cm}\includegraphics[width=0.18\linewidth]{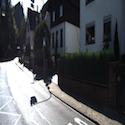}\\
			\hspace{-0.0cm}\includegraphics[width=0.18\linewidth]{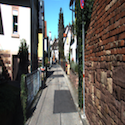}&
			\hspace{-0.05cm}\includegraphics[width=0.18\linewidth]{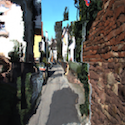}&
			\hspace{-0.05cm}\includegraphics[width=0.18\linewidth]{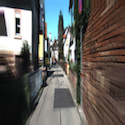}&
			\hspace{-0.05cm}\includegraphics[width=0.18\linewidth]{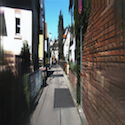}&			
			\hspace{-0.05cm}\includegraphics[width=0.18\linewidth]{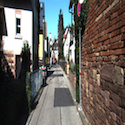}\\
			\hspace{-0.0cm}\includegraphics[width=0.18\linewidth]{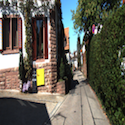}&
			\hspace{-0.05cm}\includegraphics[width=0.18\linewidth]{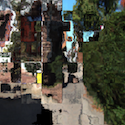}&
			\hspace{-0.05cm}\includegraphics[width=0.18\linewidth]{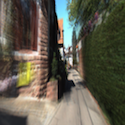}&
			\hspace{-0.05cm}\includegraphics[width=0.18\linewidth]{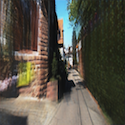}&			
			\hspace{-0.05cm}\includegraphics[width=0.18\linewidth]{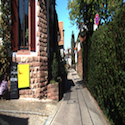}\\
			\hspace{-0.0cm}\includegraphics[width=0.18\linewidth]{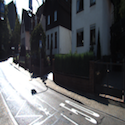}&
			\hspace{-0.05cm}\includegraphics[width=0.18\linewidth]{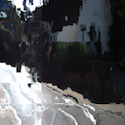}&
			\hspace{-0.05cm}\includegraphics[width=0.18\linewidth]{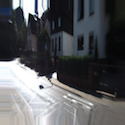}&
			\hspace{-0.05cm}\includegraphics[width=0.18\linewidth]{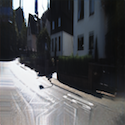}&		
			\hspace{-0.05cm}\includegraphics[width=0.18\linewidth]{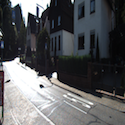}\\
			\hspace{-0.0cm}\includegraphics[width=0.18\linewidth]{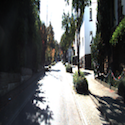}&
			\hspace{-0.05cm}\includegraphics[width=0.18\linewidth]{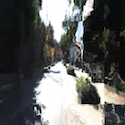}&
			\hspace{-0.05cm}\includegraphics[width=0.18\linewidth]{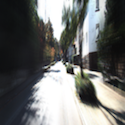}&
			\hspace{-0.05cm}\includegraphics[width=0.18\linewidth]{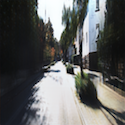}&		
			\hspace{-0.05cm}\includegraphics[width=0.18\linewidth]{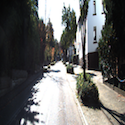}		
		\end{tabular}
	\end{small}
	\caption{{\bf Qualitative comparison of our approach with the appearance flow method of~\cite{zhou2016view} on KITTI.} While appearance flow yields artifacts, our approach, which reasons about 3D geometry, yields more realistic results.}
	\label{fig:kittires_add2}
	\vspace{-0.2cm}
\end{figure*}

\begin{figure*}[t!]
	\begin{small}
		\begin{tabular}{ccccccc}
			\hspace{-0.5cm}\begin{sideways}\hspace{0.55cm}Input view \end{sideways} &
			\hspace{-0.2cm}\includegraphics[width=0.16\linewidth]{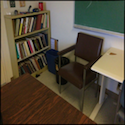}&
			\hspace{-0.2cm}\includegraphics[width=0.16\linewidth]{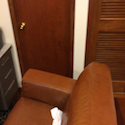}&
			\hspace{-0.2cm}\includegraphics[width=0.16\linewidth]{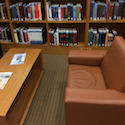}&
			\hspace{-0.2cm}\includegraphics[width=0.16\linewidth]{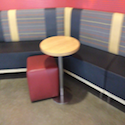}&
			\hspace{-0.2cm}\includegraphics[width=0.16\linewidth]{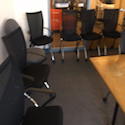}&
			\hspace{-0.2cm}\includegraphics[width=0.16\linewidth]{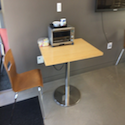}\\
			\hspace{-0.5cm}\begin{sideways}\hspace{1cm}OursGeo \end{sideways}&
			\hspace{-0.2cm}\includegraphics[width=0.16\linewidth]{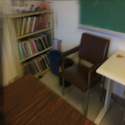}&
			\hspace{-0.2cm}\includegraphics[width=0.16\linewidth]{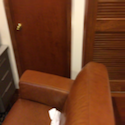}&
			\hspace{-0.2cm}\includegraphics[width=0.16\linewidth]{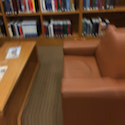}&
			\hspace{-0.2cm}\includegraphics[width=0.16\linewidth]{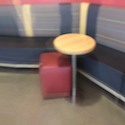}&
			\hspace{-0.2cm}\includegraphics[width=0.16\linewidth]{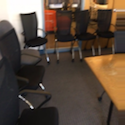}&
			\hspace{-0.2cm}\includegraphics[width=0.16\linewidth]{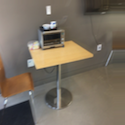}\\
			\hspace{-0.5cm}\begin{sideways}\hspace{1cm}OursFull \end{sideways}&
			\hspace{-0.2cm}\includegraphics[width=0.16\linewidth]{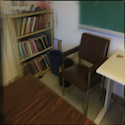}&
			\hspace{-0.2cm}\includegraphics[width=0.16\linewidth]{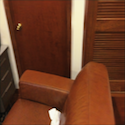}&
			\hspace{-0.2cm}\includegraphics[width=0.16\linewidth]{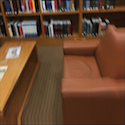}&
			\hspace{-0.2cm}\includegraphics[width=0.16\linewidth]{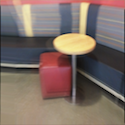}&
			\hspace{-0.2cm}\includegraphics[width=0.16\linewidth]{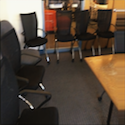}&
			\hspace{-0.2cm}\includegraphics[width=0.16\linewidth]{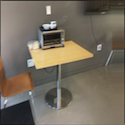}\\			
			\hspace{-0.5cm}\begin{sideways}\hspace{0.5cm}Ground-truth \end{sideways} &
			\hspace{-0.2cm}\includegraphics[width=0.16\linewidth]{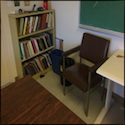}&
			\hspace{-0.2cm}\includegraphics[width=0.16\linewidth]{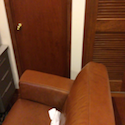}&
			\hspace{-0.2cm}\includegraphics[width=0.16\linewidth]{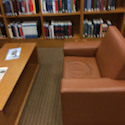}&
			\hspace{-0.2cm}\includegraphics[width=0.16\linewidth]{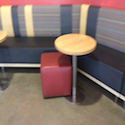}&
			\hspace{-0.2cm}\includegraphics[width=0.16\linewidth]{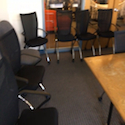}&
			\hspace{-0.2cm}\includegraphics[width=0.16\linewidth]{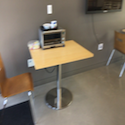}
		\end{tabular}
	\end{small}
	\caption{{\bf Qualitative results of our approach  on ScanNet.} }
	\label{fig:scannetres-v1}
\end{figure*}

We provide additional qualitative results on the KITTI dataset in Figs.~\ref{fig:kittires_add1}, and~\ref{fig:kittires_add2} 
and on the ScanNet dataset in Fig.~\ref{fig:scannetres-v1}. 
As those in the main paper, they clearly illustrate the benefits of our approach over the state-of-the-art appearance flow baseline~\cite{zhou2016view}; specifically, accounting for geometry lets us produce much more realistic novel views. Note also that our complete approach (Ours-Full), with the refinement network, typically yields sharper results than our basic framework without refinement (Ours-Geo). This can be seen, e.g., in the third to seventh rows of Fig.~\ref{fig:kittires_add1}.

In Table~\ref{tab:comparisonScannet}, we analyze the influence of the quality of the depth and normal estimates and of learning the selection maps on ScanNet. Note that, compared to Table~2 in the main paper which shows a similar analysis for KITTI, we eliminated the factor `gtNor' because it is computed from `gtDep'. In essence, the behavior is the same as for KITTI. The best results are obtained with the ground-truth depth maps, which leaves room for our method to improve as progress in depth estimation is made. More importantly, our learnt selection maps give a significant boost to our results, whether using ground-truth depth or estimated one.


\subsection{Synthesized Candidate Images}
\begin{figure*}[t!]
	\begin{tabular}{cccc}
		\hspace{0.4cm} Input image& \hspace{-0.2cm} Ground-truth novel view&  \hspace{-0.2cm} Ours-Full& Ours-Geo \\
		\hspace{0.4cm}\includegraphics[width=0.22\linewidth]{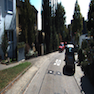} & \hspace{-0.2cm}\includegraphics[width=0.22\linewidth]{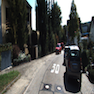}& \hspace{-0.2cm}\includegraphics[width=0.22\linewidth]{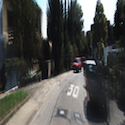}  &\hspace{-0.2cm}\includegraphics[width=0.22\linewidth]{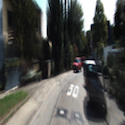}\\
		& & \hspace{-4.5cm} Synthesized images  & \\
		\hspace{0.4cm}\includegraphics[width=0.22\linewidth]{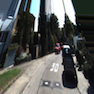}& \hspace{-0.2cm}\includegraphics[width=0.22\linewidth]{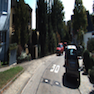}&
		\hspace{-0.2cm}\includegraphics[width=0.22\linewidth]{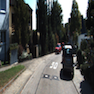}&
		\hspace{-0.2cm}\includegraphics[width=0.22\linewidth]{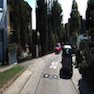}\\
		\hspace{0.4cm}\includegraphics[width=0.22\linewidth]{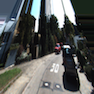}& \hspace{-0.2cm}\includegraphics[width=0.22\linewidth]{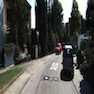}&
		\hspace{-0.2cm}\includegraphics[width=0.22\linewidth]{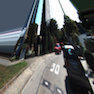}&
		\hspace{-0.2cm}\includegraphics[width=0.22\linewidth]{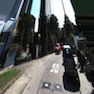}\\
		\hspace{0.4cm}\includegraphics[width=0.22\linewidth]{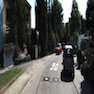}& \hspace{-0.2cm}\includegraphics[width=0.22\linewidth]{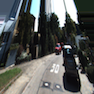}&
		\hspace{-0.2cm}\includegraphics[width=0.22\linewidth]{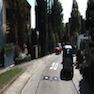}&
		\hspace{-0.2cm}\includegraphics[width=0.22\linewidth]{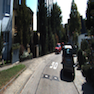}\\
		\hspace{0.4cm} \includegraphics[width=0.22\linewidth]{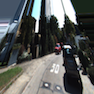}& \hspace{-0.2cm}\includegraphics[width=0.22\linewidth]{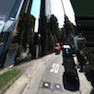}&
		\hspace{-0.2cm}\includegraphics[width=0.22\linewidth]{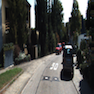}&
		\hspace{-0.2cm}\includegraphics[width=0.22\linewidth]{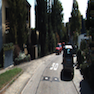}
	\end{tabular}
	\vspace{0.05cm}
	\caption{{\bf Synthesized images from the 16 estimated homographies.} In the top row, we show the input image, the ground-truth novel view and the results of our complete model (Ours-Full) and of our model without refinement (Ours-Geo). The remaining images correspond to images synthesized with our estimated homographies. Note that different homographies correctly account for the motion of different regions between the input and novel view. For instance, the top-left image models the motion of the road, while the bottom-right one accounts for the motion of the buildings.
	}
	\label{fig:16synimgs}
\end{figure*}

In Fig.~\ref{fig:16synimgs}, we show the synthesized images obtained from our $m=16$ predicted homographies for one input image. When compared with the ground-truth novel view, we can see that different homographies account for the motion of different regions in the image. For instance, the homography corresponding to the top-left image accounts for the motion of the road. By contrast, the homography corresponding to the bottom-right image models the motion of the buildings. Correctly combining these images then allows us to obtain a realistic novel view, as shown in the top row of Fig.~\ref{fig:16synimgs}.

\begin{table}
	\centering
	\small
	\begin{tabular}{c|c|c|c|c|c}
		\hline
		gtDep&estDep&estNor&Seed&SelMap&$\ell_1$\\
		\hline		
		\cmark&\xmark&\cmark&\cmark&\xmark&0.174\\
		\hline
		\cmark&\xmark&\cmark&\xmark&\cmark&0.159\\
		\hline
		\xmark&\cmark&\cmark&\cmark&\xmark&0.184\\
		\hline
		\xmark&\cmark&\cmark&\xmark&\cmark&0.167\\
		\hline
	\end{tabular}
	\caption{{\bf Influence of the quality of the depth and normal estimates and of learning the selection maps on  ScanNet.} From left to right: gtDep denotes the ground-truth depth; estDep and estNor denote the estimated depth and normal, respectively; Seed and SelMap denote the hard-segmentations corresponding to the seed regions and the selection map obtained with our selection network, respectively.
	}
	\label{tab:comparisonScannet}
\end{table}

\subsection{Selection Maps}
In Fig.~\ref{fig:selectionMapimgs}, we provide additional results from our selection network. While our seed regions typically cover only parts of the road, trees, sky, and buildings, our predicted selection maps can extend them to larger planar and semantically-coherent regions. 
\begin{figure*}[t!]
	\begin{tabular}{cccccc}
		\hspace{0.4cm} & \hspace{-0.2cm}Input image &&&\hspace{-0.2cm}Input Image \hspace{-0.2cm} &\\
		&\hspace{0.4cm}\hspace{-0.2cm}\includegraphics[width=0.15\linewidth]{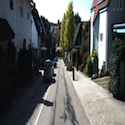}&& & \hspace{-0.2cm}\includegraphics[width=0.15\linewidth]{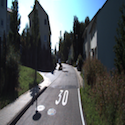}  &\hspace{-0.2cm} \vspace{0.2cm}\\
		Seed Region& Selection Map &  Overlay Image & Seed Region& Selection Map &  Overlay Image\\
		\hspace{0.4cm}\includegraphics[width=0.15\linewidth]{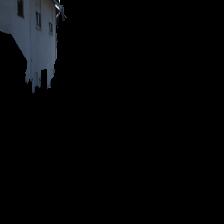}& \hspace{-0.2cm}\includegraphics[width=0.15\linewidth]{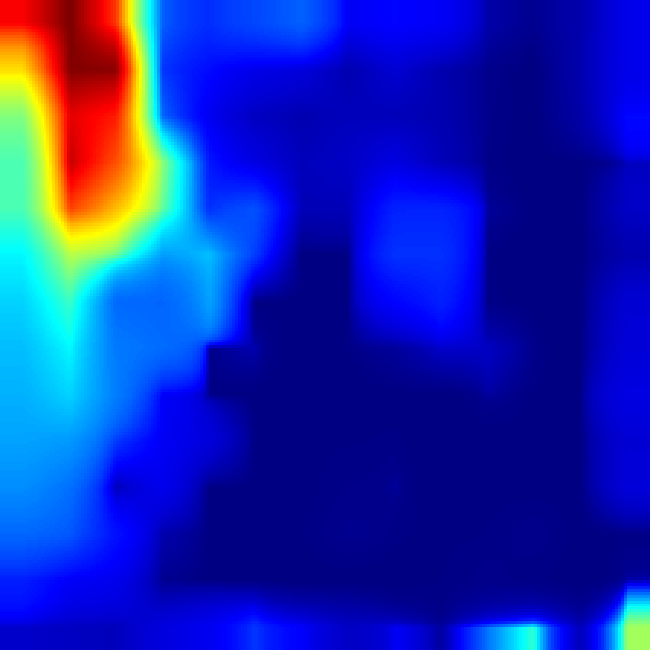}&
		\hspace{-0.2cm}\includegraphics[width=0.15\linewidth]{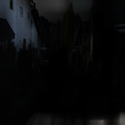}&
		\hspace{-0.2cm}\includegraphics[width=0.15\linewidth]{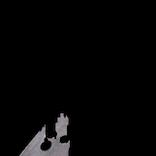}&
		\hspace{-0.2cm}\includegraphics[width=0.15\linewidth]{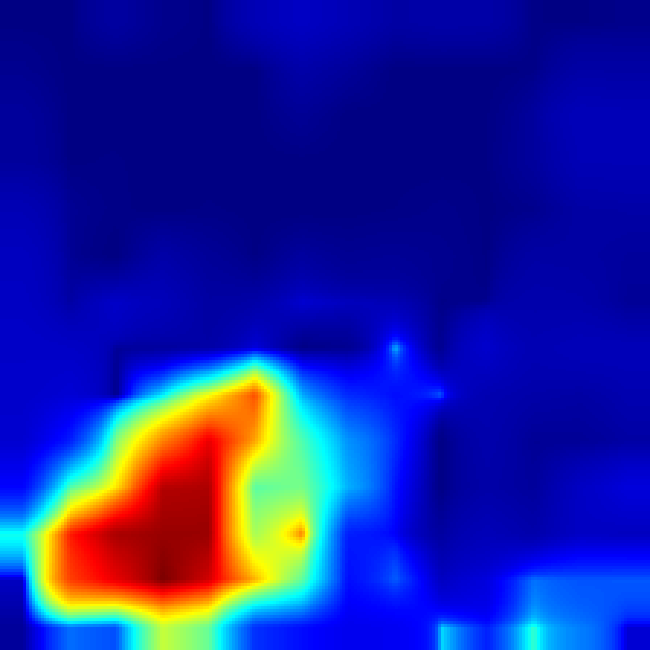}&
		\hspace{-0.2cm}\includegraphics[width=0.15\linewidth]{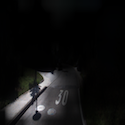}\\
		\hspace{0.4cm}\includegraphics[width=0.15\linewidth]{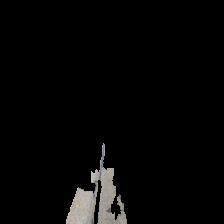}& \hspace{-0.2cm}\includegraphics[width=0.15\linewidth]{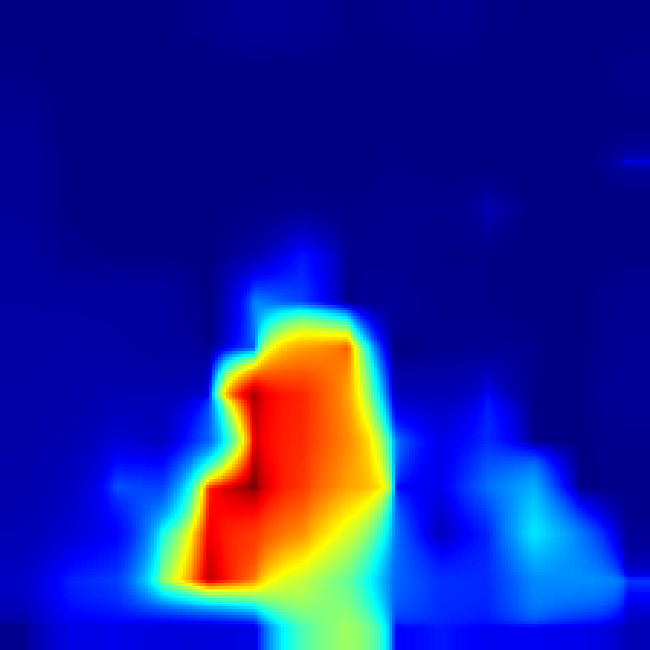}&
		\hspace{-0.2cm}\includegraphics[width=0.15\linewidth]{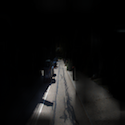}&
		\hspace{-0.2cm}\includegraphics[width=0.15\linewidth]{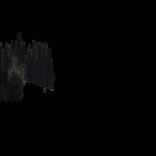}&
		\hspace{-0.2cm}\includegraphics[width=0.15\linewidth]{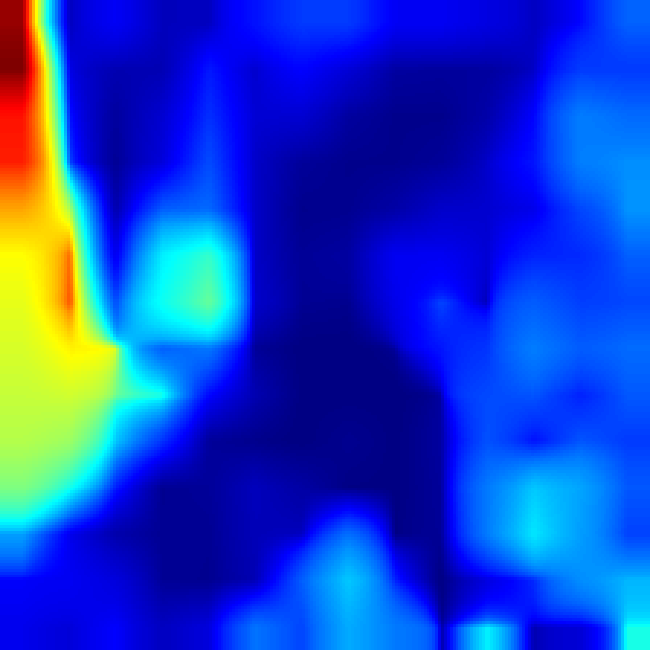}&
		\hspace{-0.2cm}\includegraphics[width=0.15\linewidth]{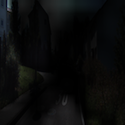}\\
		\hspace{0.4cm}\includegraphics[width=0.15\linewidth]{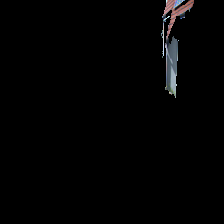}& \hspace{-0.2cm}\includegraphics[width=0.15\linewidth]{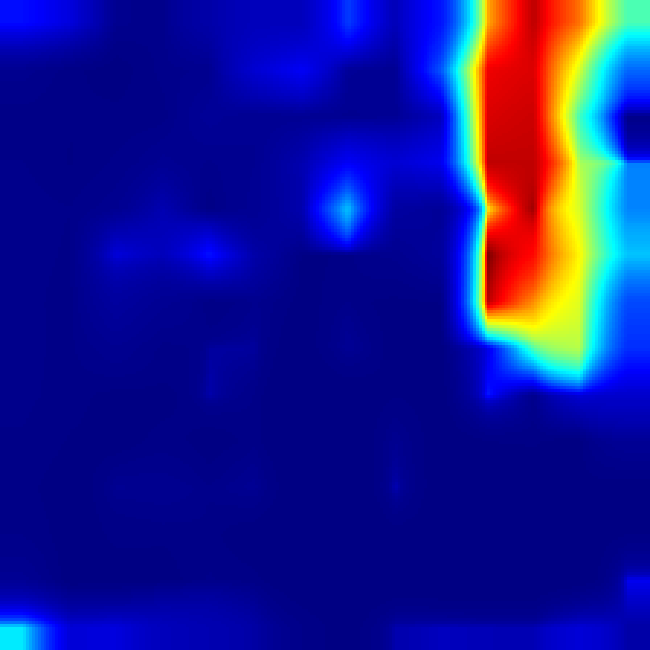}&
		\hspace{-0.2cm}\includegraphics[width=0.15\linewidth]{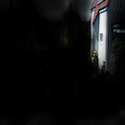}&
		\hspace{-0.2cm}\includegraphics[width=0.15\linewidth]{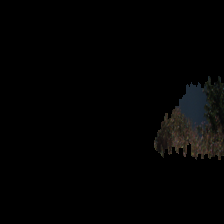}&
		\hspace{-0.2cm}\includegraphics[width=0.15\linewidth]{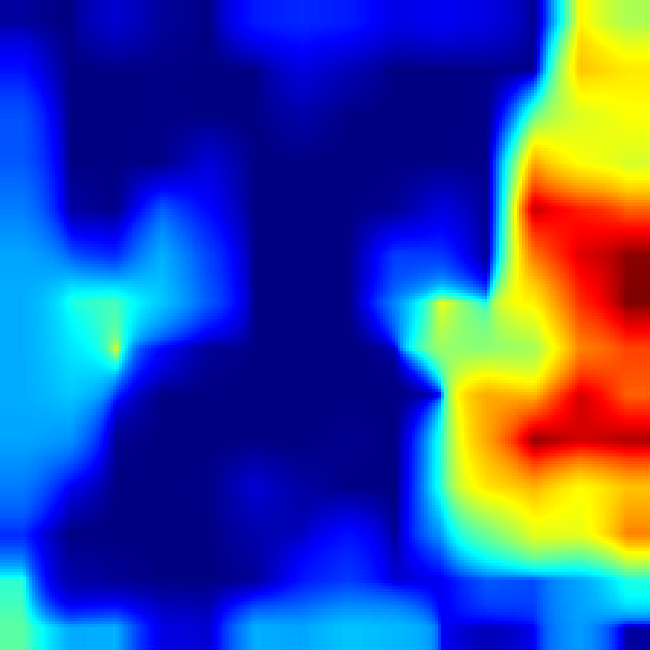}&
		\hspace{-0.2cm}\includegraphics[width=0.15\linewidth]{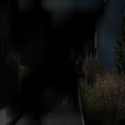}\\
		\hspace{0.4cm} \includegraphics[width=0.15\linewidth]{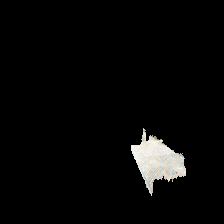}& \hspace{-0.2cm}\includegraphics[width=0.15\linewidth]{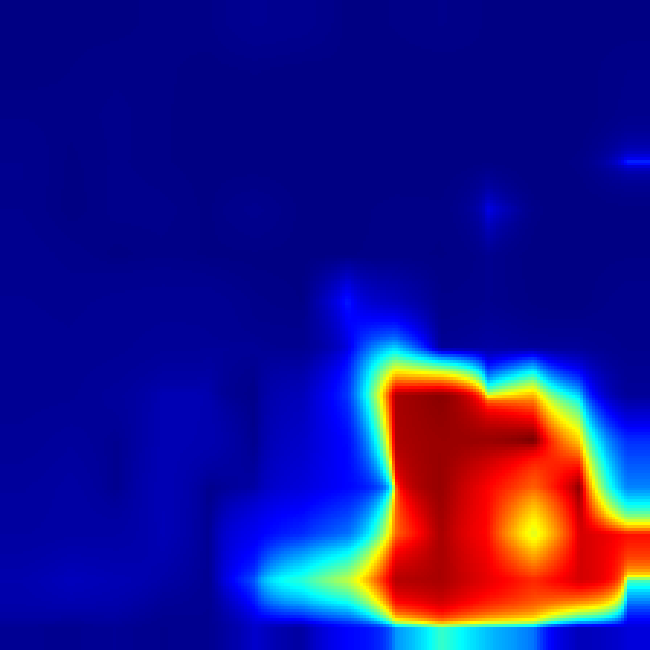}&
		\hspace{-0.2cm}\includegraphics[width=0.15\linewidth]{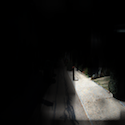}&
		\hspace{-0.2cm}\includegraphics[width=0.15\linewidth]{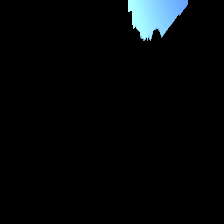}&
		\hspace{-0.2cm}\includegraphics[width=0.15\linewidth]{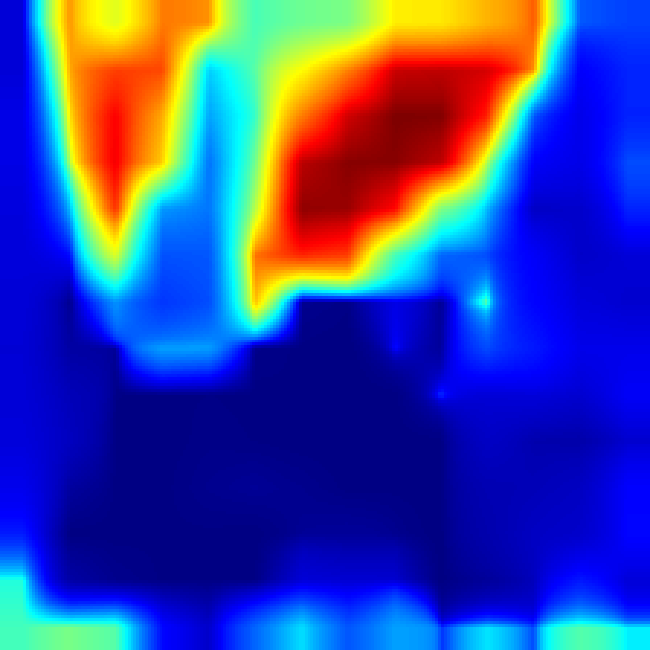}&
		\hspace{-0.2cm}\includegraphics[width=0.15\linewidth]{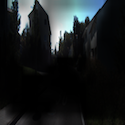}
	\end{tabular}
	\vspace{0.05cm}
	\caption{{\bf Sample seed regions and predicted selection maps in the input view.} From left to right: seed region, predicted selection map and predicted selection map overlaid on the input image. Red indicates a high likelihood for a pixel to belong to the plane defined by the seed region and blue a low likelihood.}
	\label{fig:selectionMapimgs}
\end{figure*}

\subsection{Depth and Normal prediction}
\begin{figure*}[h]
	\vspace{-0.3cm}
	\begin{small}
		\begin{tabular}{cccc}
			Image& GT-depth & Est-depth & Est-normal\\ 
			\includegraphics[width=0.23\linewidth]{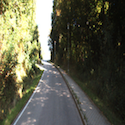}&
			\includegraphics[width=0.23\linewidth]{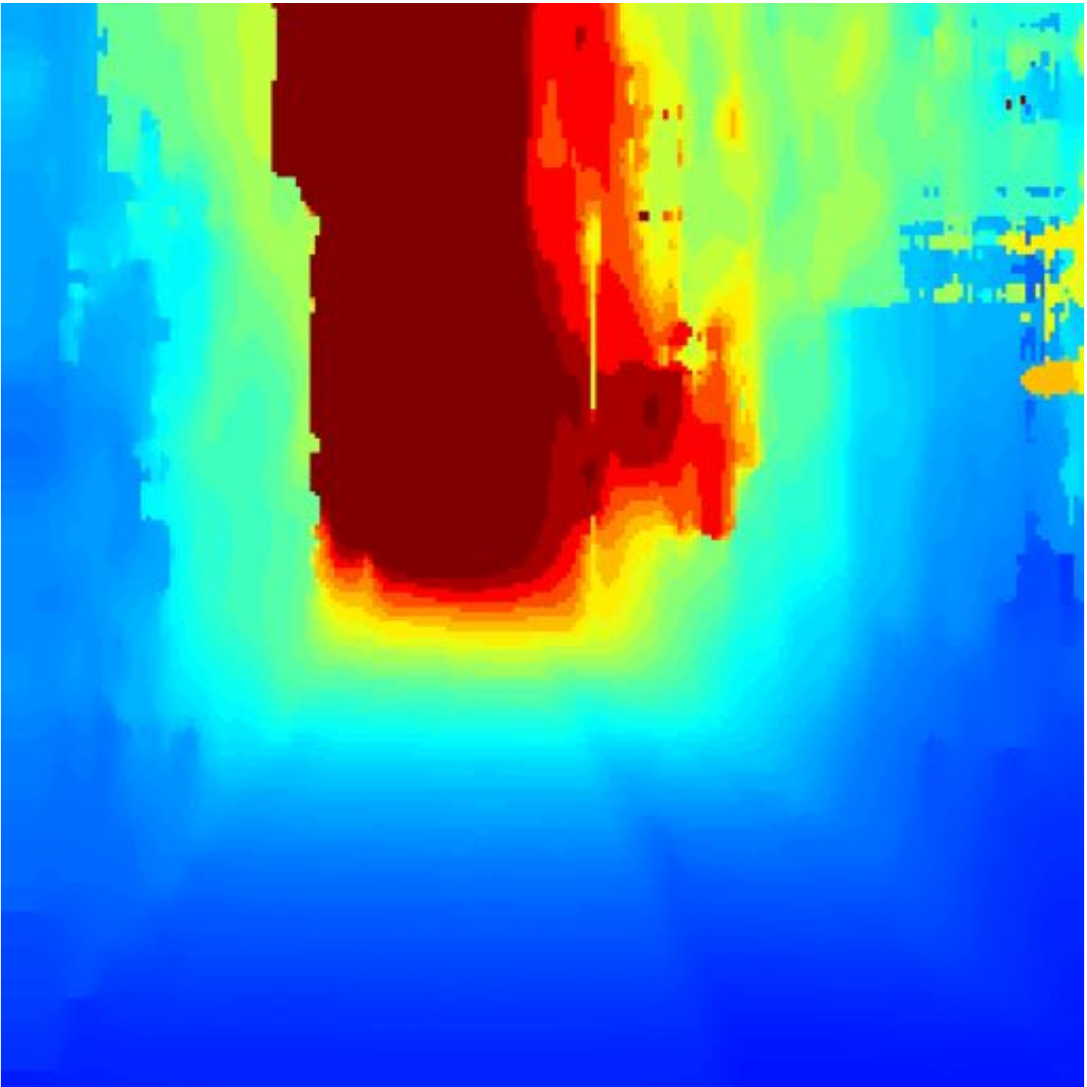}&
			\includegraphics[width=0.23\linewidth]{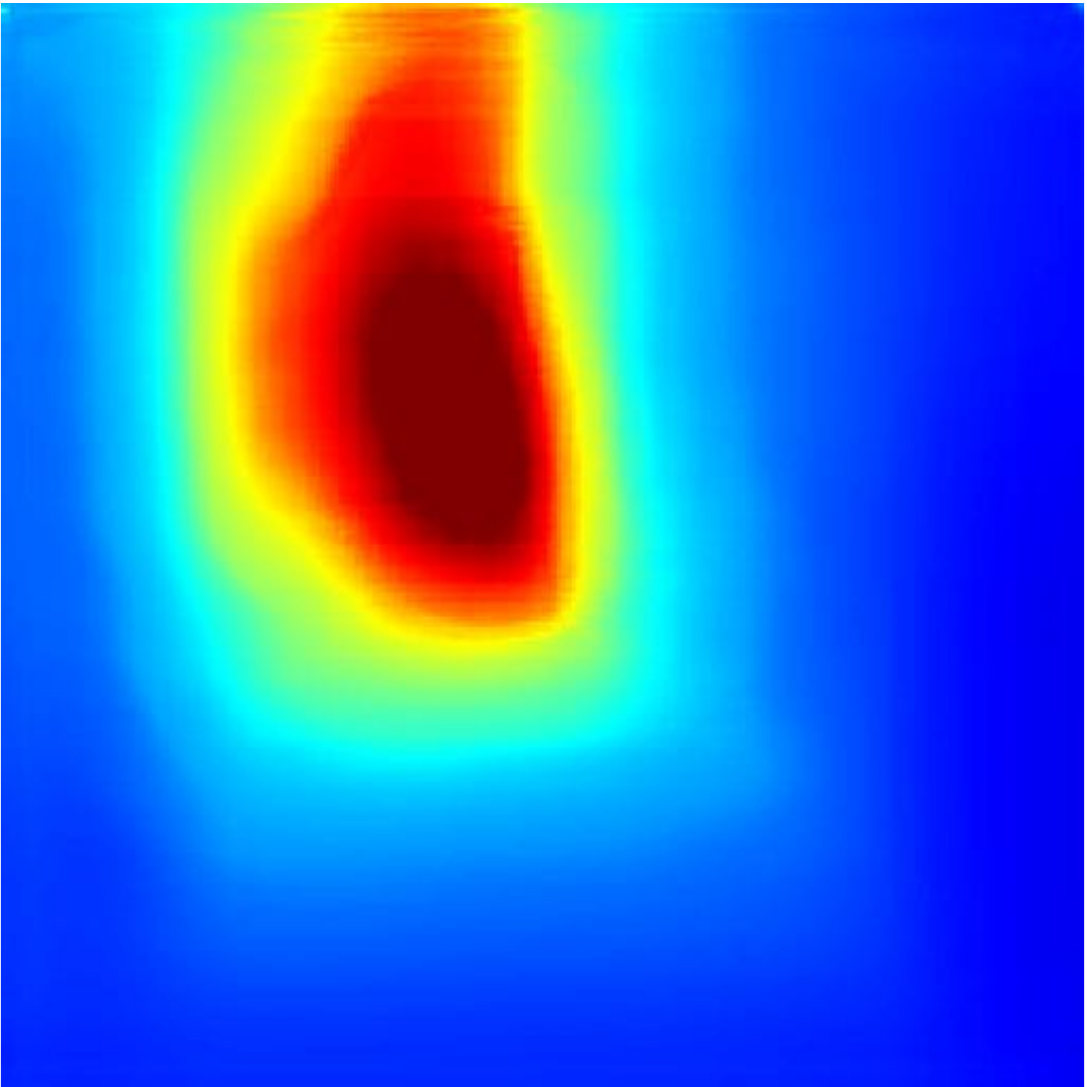}&
			\includegraphics[width=0.23\linewidth]{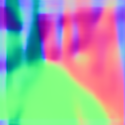}\\
			\includegraphics[width=0.23\linewidth]{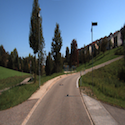}&
			\includegraphics[width=0.23\linewidth]{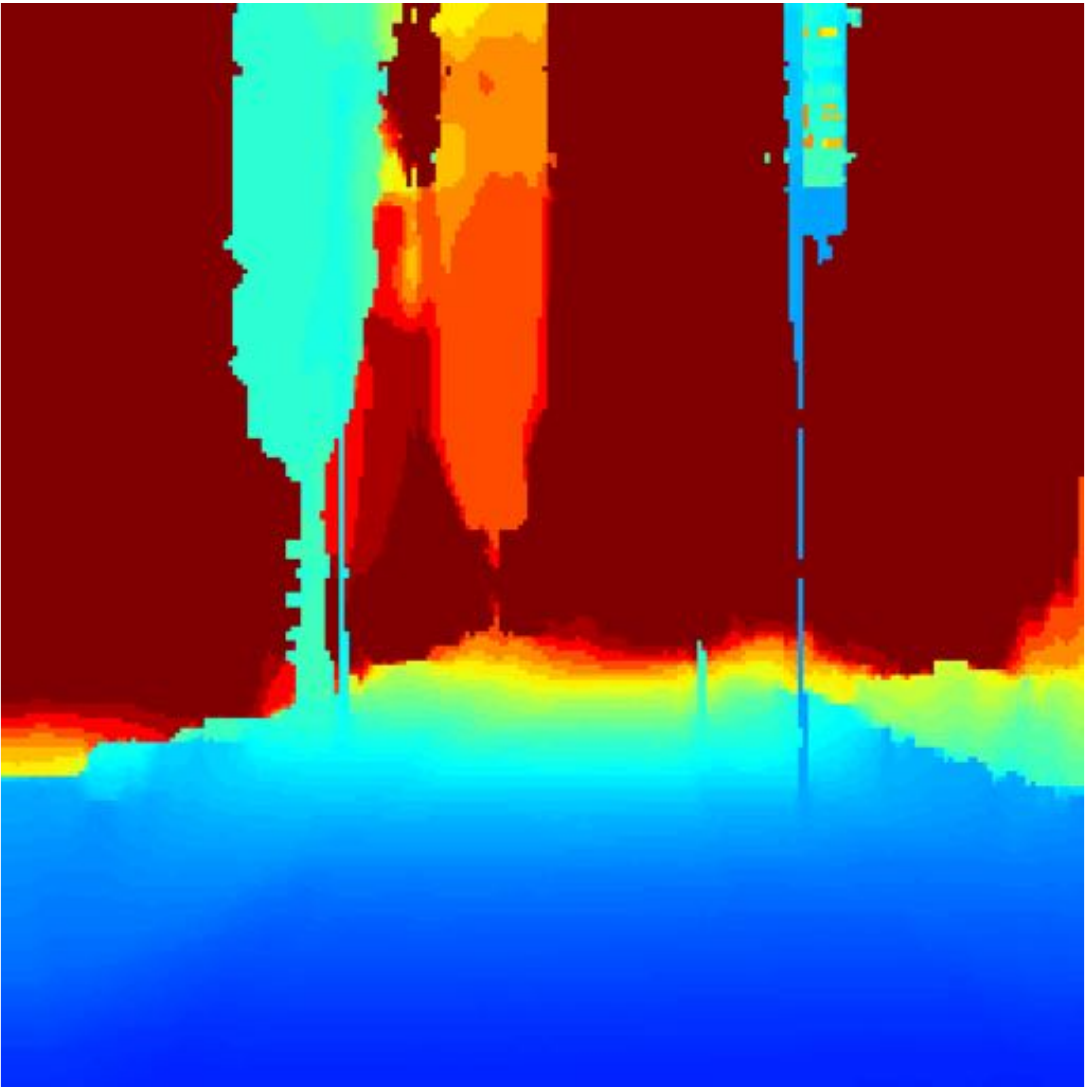}&
			\includegraphics[width=0.23\linewidth]{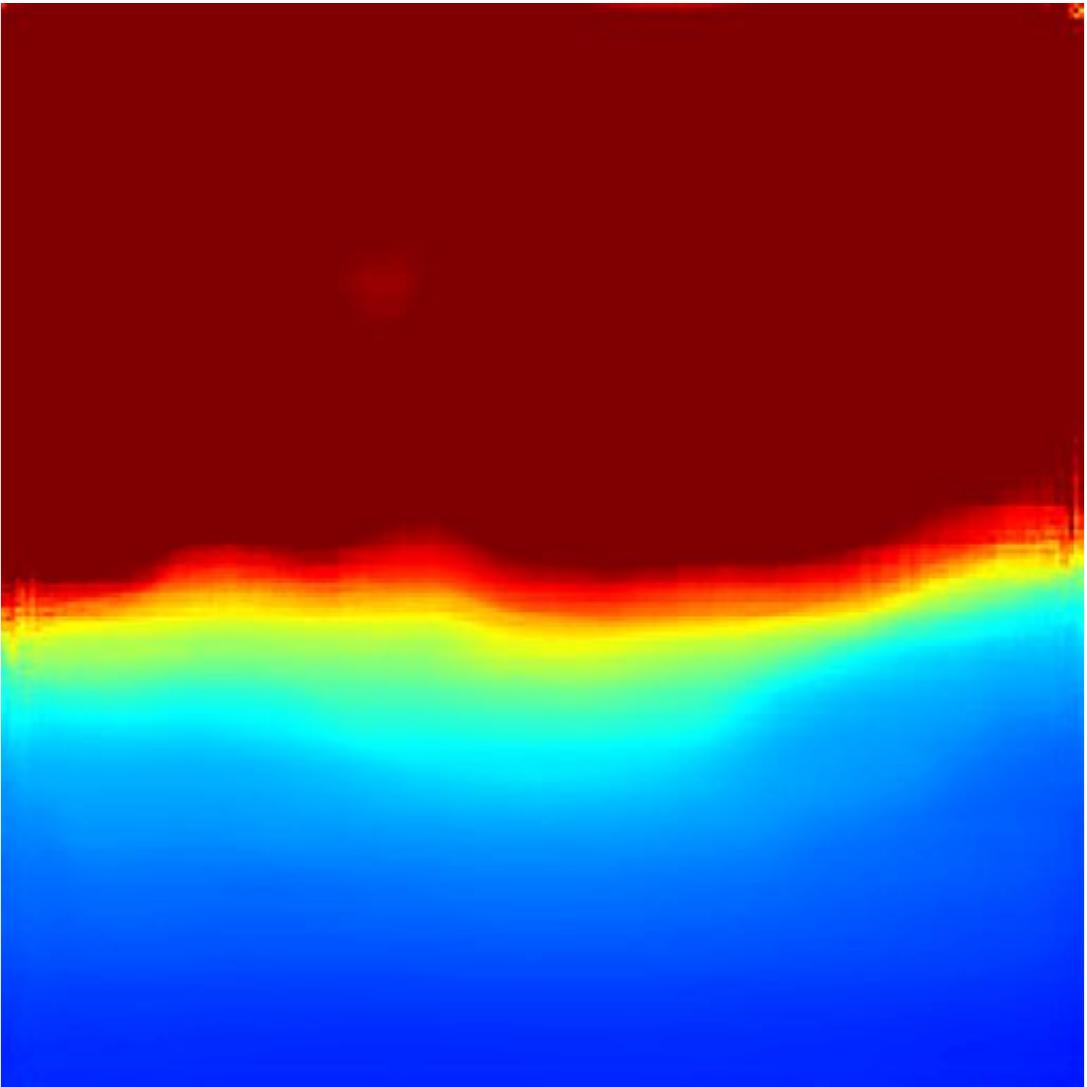}&
			\includegraphics[width=0.23\linewidth]{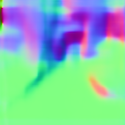}\\
			\includegraphics[width=0.23\linewidth]{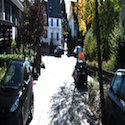}&
			\includegraphics[width=0.23\linewidth]{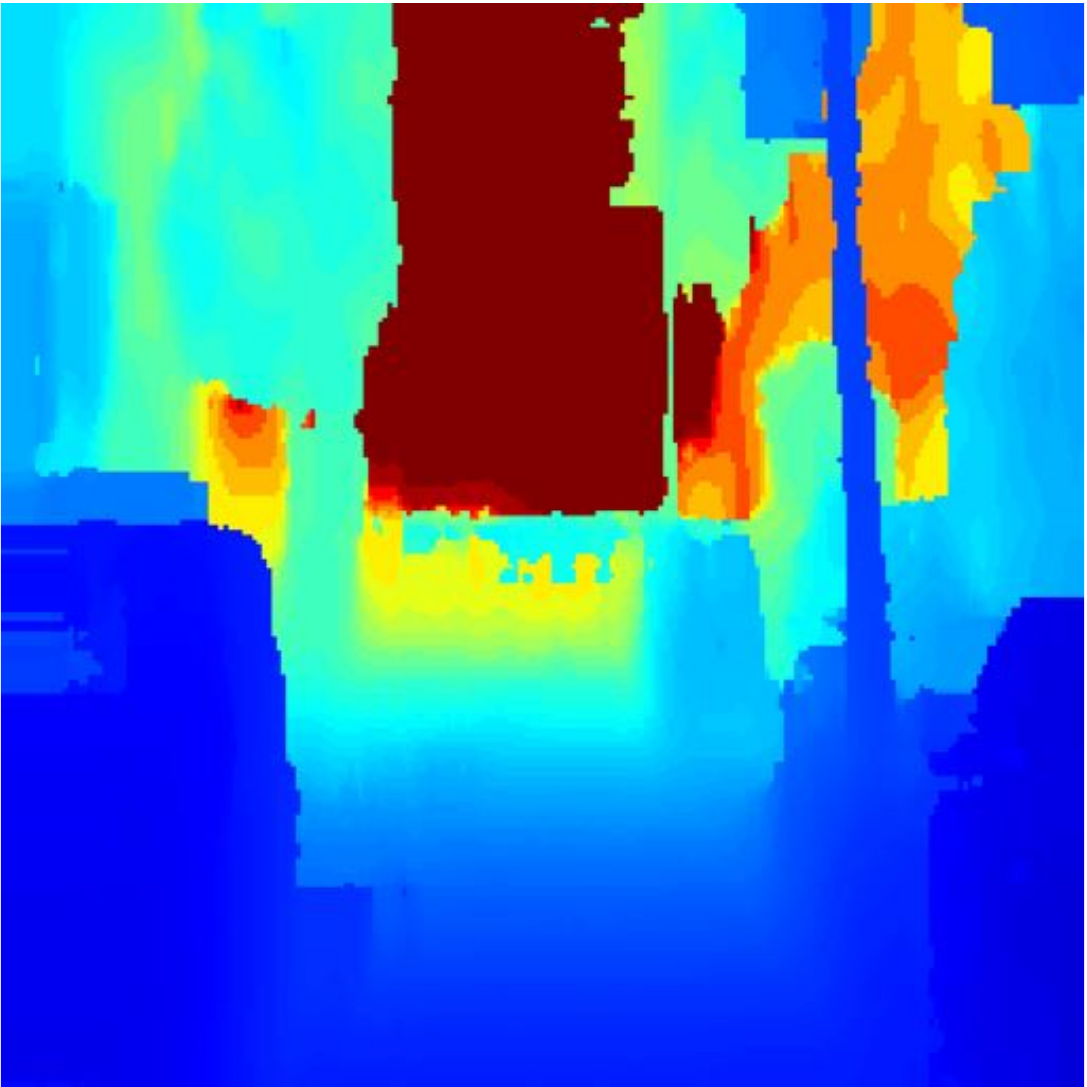}&
			\includegraphics[width=0.23\linewidth]{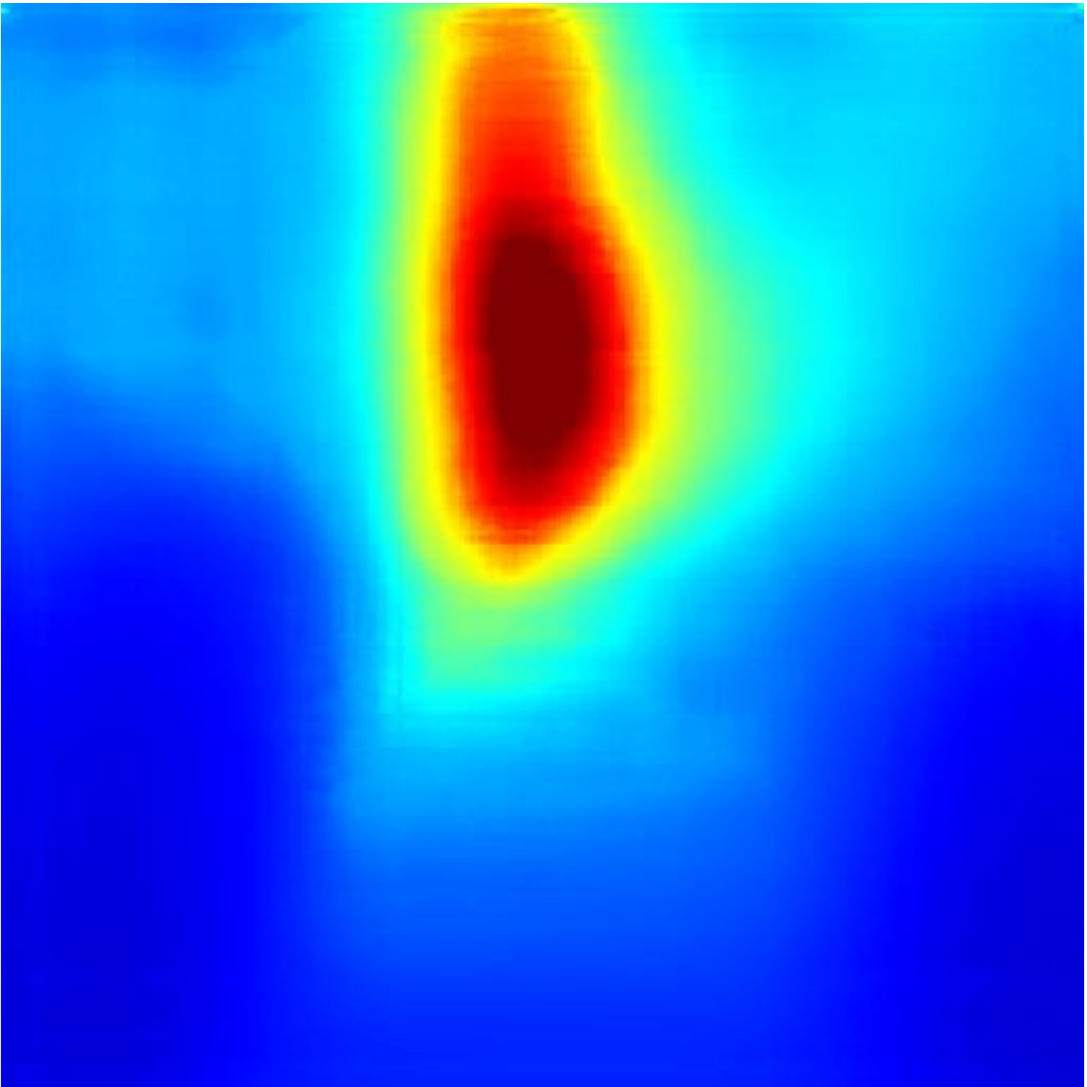}&
			\includegraphics[width=0.23\linewidth]{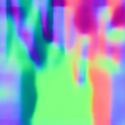}\\
			\includegraphics[width=0.23\linewidth]{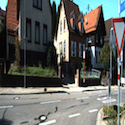}&
			\includegraphics[width=0.23\linewidth]{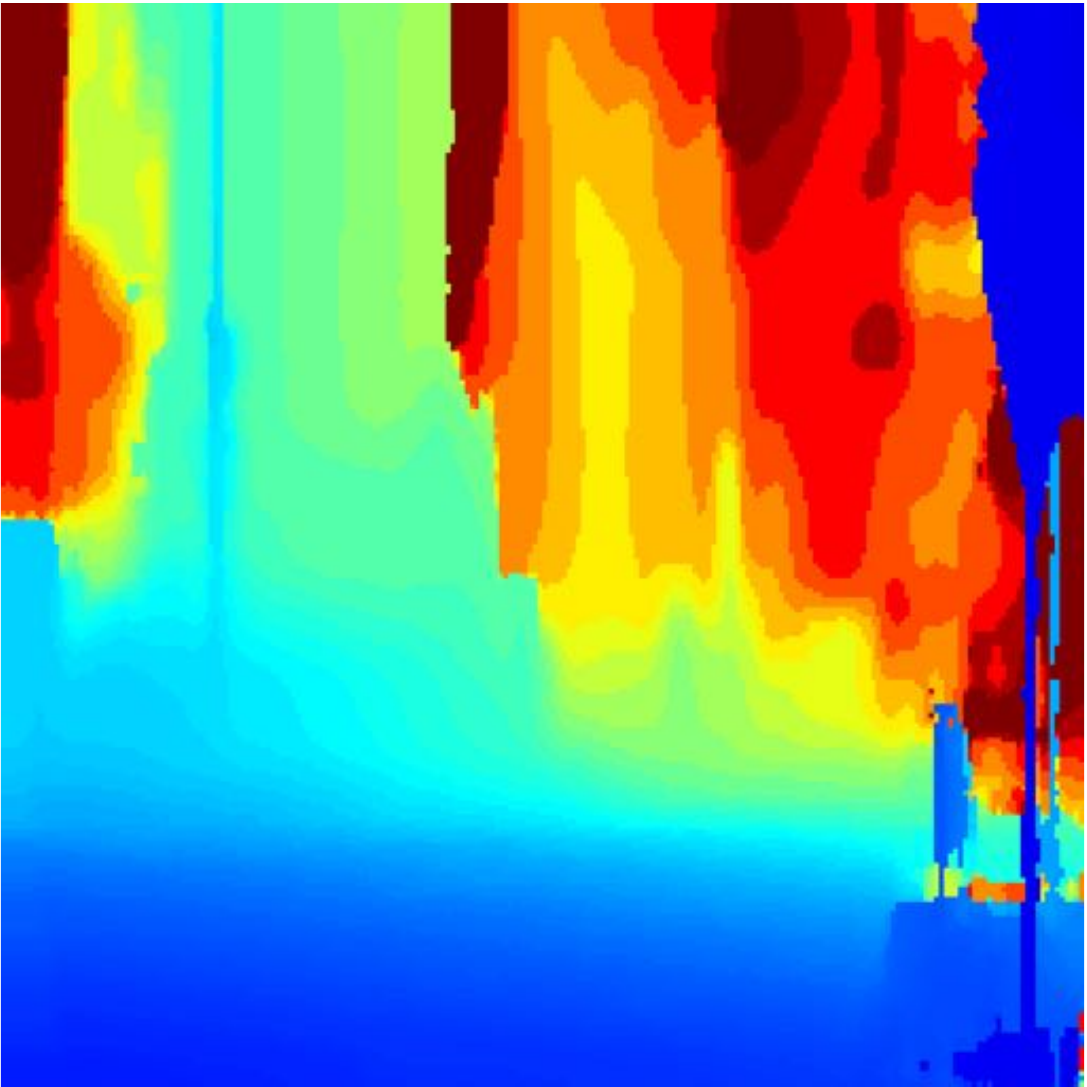}&
			\includegraphics[width=0.23\linewidth]{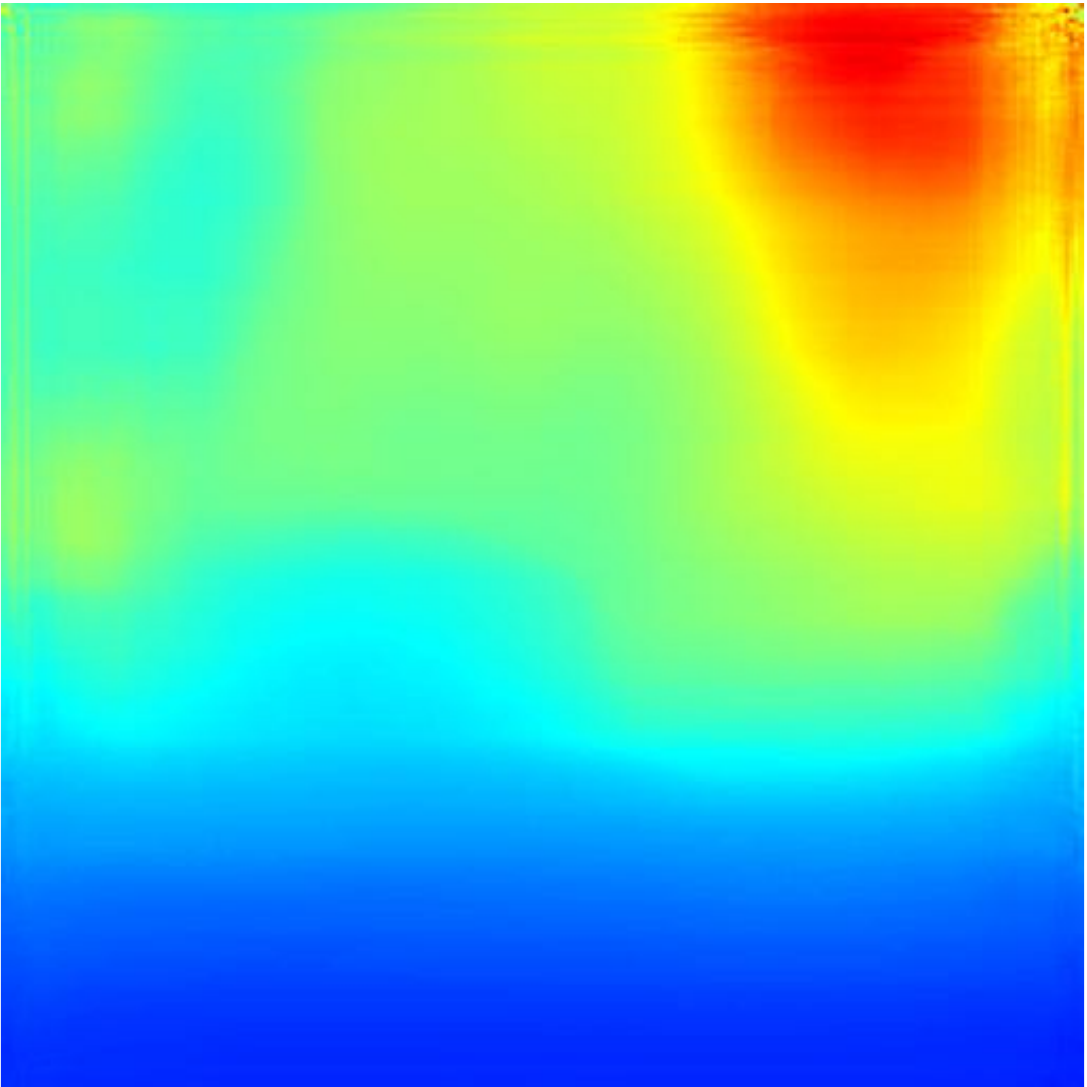}&
			\includegraphics[width=0.23\linewidth]{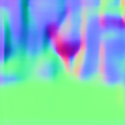}\\
			\includegraphics[width=0.23\linewidth]{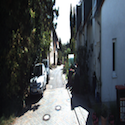}&
			\includegraphics[width=0.23\linewidth]{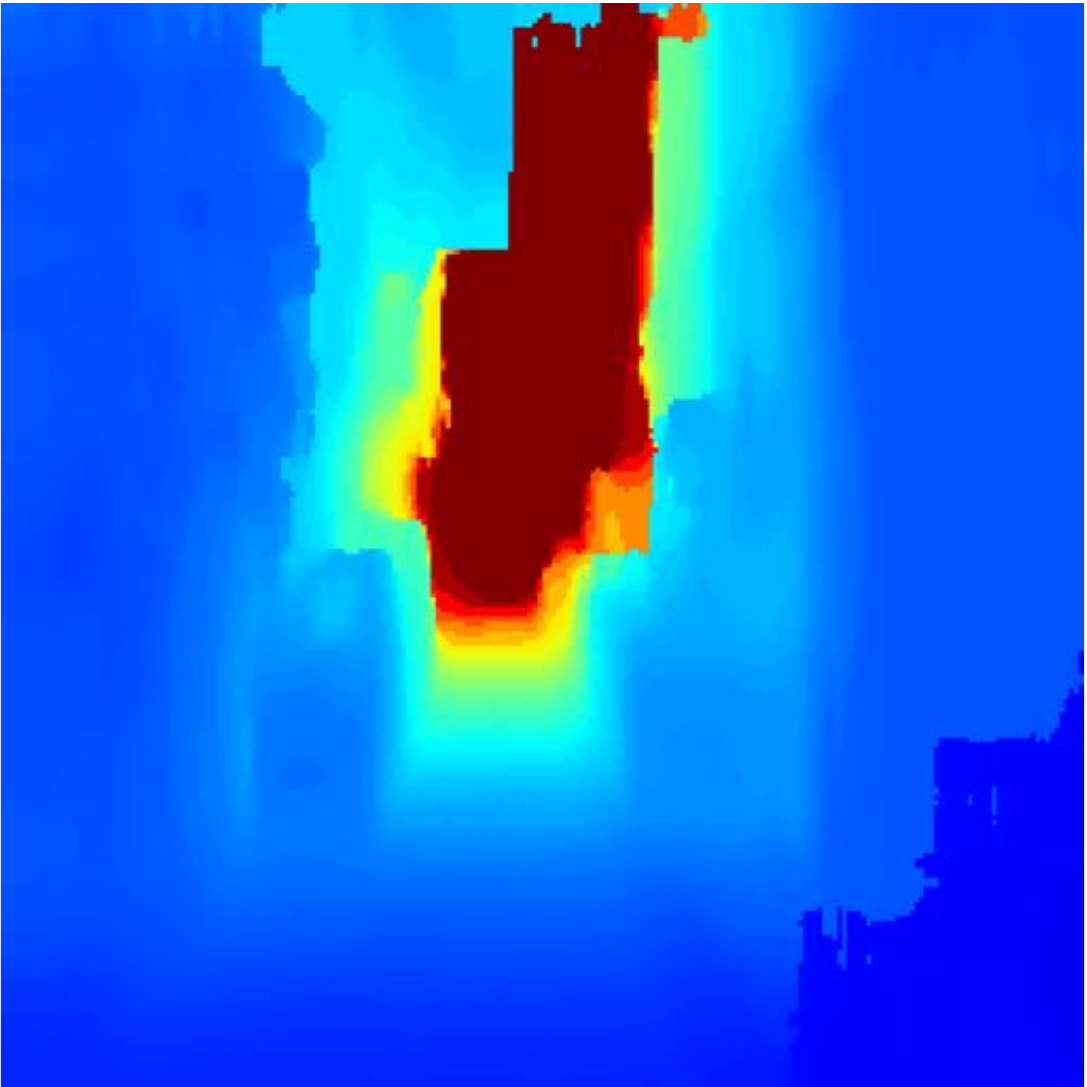}&
			\includegraphics[width=0.23\linewidth]{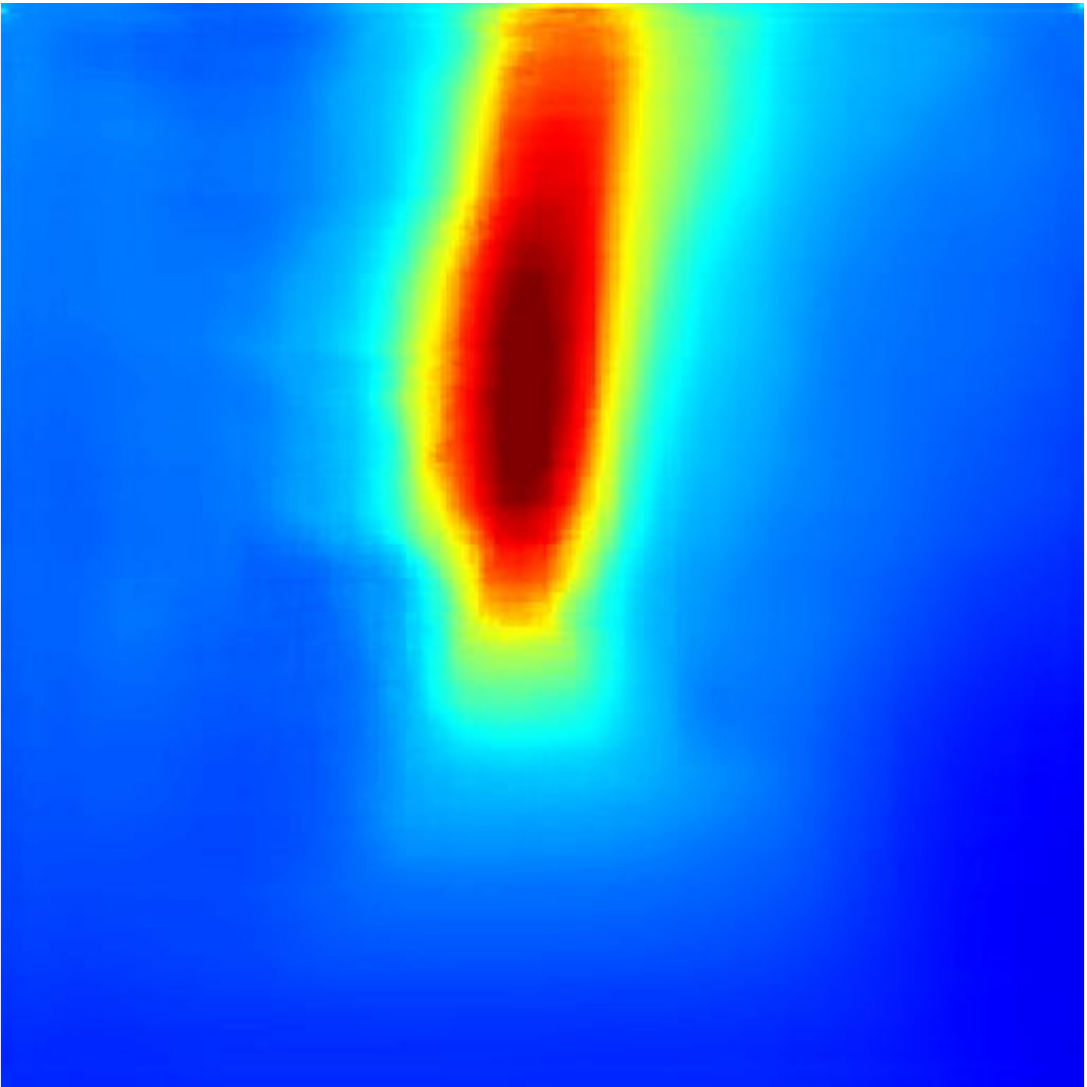}&
			\includegraphics[width=0.23\linewidth]{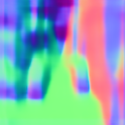}\\		
		\end{tabular}
	\end{small}
	\vspace{-0.2cm}
	\caption{{\bf Visualization of the estimated depth and normal for KITTI.} Color indicates depth (red is far, blure is close).}
	\label{fig:depnorVisualization}
	\vspace{-0.5cm}
\end{figure*}

In Fig.~\ref{fig:depnorVisualization}, we provide the visualisation of the estimated depth and normal map from our network for sampled images from KITTI test set. It shows that our estimation can well capture the scene structure compared with the ground truth.

\subsection{Failure Cases}
\begin{figure*}[h]
	\vspace{-0.3cm}
	\begin{small}
		\begin{tabular}{cccc}
			input view& Ours-Geo &Ours-Full &Ground-truth\\ 
			\includegraphics[width=0.23\linewidth]{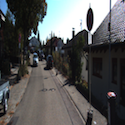}&
			\includegraphics[width=0.23\linewidth]{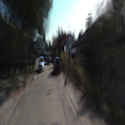}&
			\includegraphics[width=0.23\linewidth]{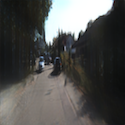}&
			\includegraphics[width=0.23\linewidth]{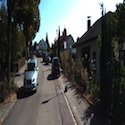}\\
			\includegraphics[width=0.23\linewidth]{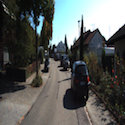}&
			\includegraphics[width=0.23\linewidth]{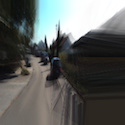}&
			\includegraphics[width=0.23\linewidth]{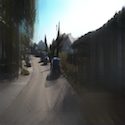}&
			\includegraphics[width=0.23\linewidth]{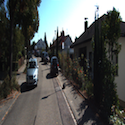}\\
			\includegraphics[width=0.23\linewidth]{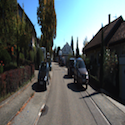}&
			\includegraphics[width=0.23\linewidth]{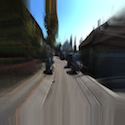}&
			\includegraphics[width=0.23\linewidth]{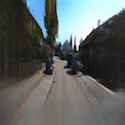}&
			\includegraphics[width=0.23\linewidth]{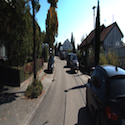}
		\end{tabular}
	\end{small}
	\vspace{-0.2cm}
	\caption{{\bf Failure cases of our approach on KITTI.} Typical failures correspond to moving objects, or hallucination of large portions of the image (e.g., due to backward motion), in which case our approach tends to generate background instead of foreground objects.}
	\label{fig:failureexps}
	\vspace{-0.5cm}
\end{figure*}

In Fig.~\ref{fig:failureexps}, we show typical failure cases of our approach. The failure cases are mainly due to i) moving objects, whose locations cannot be explained by camera motion (see the first row); 2) the need to hallucinate large portions of the image (e.g., because of backward motion), in which case our method tends to generate background and miss foreground objects (see the last two examples).

\end{document}